\documentclass{article}

\usepackage[preprint]{neurips_2024}

\usepackage[utf8]{inputenc} % allow utf-8 input
\usepackage[T1]{fontenc}    % use 8-bit T1 fonts
\usepackage{hyperref}       % hyperlinks
\usepackage{url}            % simple URL typesetting
\usepackage{booktabs}       % professional-quality tables
\usepackage{amsfonts}       % blackboard math symbols
\usepackage{nicefrac}       % compact symbols for 1/2, etc.
\usepackage{microtype}      % microtypography
\usepackage{xcolor}         % colors
\usepackage{graphicx}
\usepackage{subfig}
\usepackage{amsmath}
\usepackage{amssymb}
\usepackage{mathtools}
\usepackage{amsthm}
\usepackage{multirow}
\usepackage{multicol}
\usepackage{colortbl}  %彩色表格需要加载的宏包
\usepackage{xcolor}
\usepackage{makecell}
\usepackage{wrapfig}
\usepackage{ulem}
\usepackage{bbding}
\makeatletter
  \newcommand\figcaption{\def\@captype{figure}\caption}
  \newcommand\tabcaption{\def\@captype{table}\caption}
\makeatother

\usepackage{algorithm}
\usepackage[noend]{algpseudocode}

\title{Reconstruct the Pruned Model without Any Retraining}

\author{%
  Pingjie Wang,  Ziqing Fan,  Shengchao Hu,  Zhe Chen\\
  \textbf{Yanfeng Wang,  Yu Wang\thanks{Corresponding author.}} \\
  Shanghai Jiao Tong University \\
  Shanghai Artificial Intelligence Laboratory \\ 
  \texttt{\{pingjiewang, zqfan\_knight, charles-hu, chenzhe2018\}@sjtu.edu.cn} \\
  \texttt{\{wangyanfeng622, yuwangsjtu\}@sjtu.edu.cn} \\
}

\begin{document}

\maketitle

\begin{abstract}
  %Structured pruning is a promising hardware-friendly compression technique for large language models (LLMs), which is expected to be retraining-free to avoid the enormous retraining cost. This retraining-free approach typically involves (\romannumeral1) \textit{pruning criteria} to define the architecture and (\romannumeral2) \textit{distortion reconstruction} to restore performance. However, existing methods often focus heavily on pruning criteria, with reconstruction techniques tailored to specific modules or criteria, resulting in limited generalizability. To address this issue, we introduce a Linear Interpolation-based Adaptive Reconstruction (LIAR) framework, which is both efficient and effective. It does not require back-propagation or retraining and is compatible with various pruning criteria and modules. LIAR minimizes reconstruction error by applying linear interpolation to the preserved weights, thereby reconstructing the pruned output. Our evaluations on benchmarks such as GLUE, SQuAD, WikiText, and common sense reasoning demonstrate that LIAR enables a BERT model to maintain 98\% accuracy despite the removal of 50\% of its parameters and achieves top performance for LLaMA in just a few minutes.
Structured pruning is a promising hardware-friendly compression technique for large language models (LLMs), which is expected to be retraining-free to avoid the enormous retraining cost. This retraining-free paradigm involves (\romannumeral1) \textit{pruning criteria} to define the architecture and (\romannumeral2) \textit{distortion reconstruction} to restore performance. However, existing methods often emphasize pruning criteria while using reconstruction techniques that are specific to certain modules or criteria, resulting in limited generalizability. To address this, we introduce the \textbf{L}inear \textbf{I}nterpolation-based \textbf{A}daptive \textbf{R}econstruction (LIAR) framework, which is both efficient and effective. LIAR does not require back-propagation or retraining and is compatible with various pruning criteria and modules. By applying linear interpolation to the preserved weights, LIAR minimizes reconstruction error and effectively reconstructs the pruned output. Our evaluations on benchmarks such as GLUE, SQuAD, WikiText, and common sense reasoning show that LIAR enables a BERT model to maintain 98\% accuracy even after removing 50\% of its parameters and achieves top performance for LLaMA in just a few minutes.

% Our code will be available after this paper is accepted.

\end{abstract}

\section{Introduction}
\label{sec:intro}

Large language models (LLMs) have attained significant achievements towards various downstream tasks in recent years \cite{zeng2022glm,workshop2022bloom,chowdhery2023palm,zhang2205opt}. However, despite the substantial progress, the deployment of LLMs is constrained by the high parameter counts and considerable computational overhead \cite{gupta2022compression}. Retraining-based structured pruning \cite{xia2022structured,tao2023structured} is one of the compression techniques to address this issue. By removing a whole group of weights from the original model, such methods can reduce the inference latency and memory storage without requiring any external hardware acceleration support \cite{neill2020overview}. However, such a strategy requires a full dataset to retrain the pruned model, resulting in significant computational overhead (e.g.~$\sim$33 hours for BERT~\cite{xia2022structured}) and extensive engineering efforts for hyper-parameter tuning and complex deployment \cite{sun2020mobilebert,jiao2019tinybert}. These requirements render the approach impractical for real-world applications, especially for LLMs.

% The conventional pruning paradigm falls into two steps: redundancy discovery and performance restoration \cite{hou2020dynabert,mccarley2019structured}. Initially, each module of the well-trained model is scored based on a specific criterion, then sorted in ascending order of importance. The top-k modules are directly removed \cite{lecun1989optimal,hassibi1993optimal}, and finally, the pruned model and/or configuration is jointly retrained to regain performance. We call this pipeline the ``criterion+retrain" paradigm. 

The Retraining-free pruning paradigm is proposed to reduce the enormous retraining consumption, which falls into two stages: 1) \textit{pruning criteria} and 2) \textit{distortion reconstruction}. For the first stage, each module of the well-trained model is scored based on a specific criterion to identify and prune redundant components~\cite{lecun1989optimal,hassibi1993optimal}. After that, the distorted output is reconstructed by the subsequent reconstruction stage. 
Compared to retraining-based approaches, the unique value of such a paradigm is its ability to regain performance without any training and only requires a small calibration dataset. Consequently, it is highly efficient (e.g., several minutes) and well-suited for compressing LLMs.

% In contrast, retraining-free pruning methodologies are fundamentally grounded in the “criterion+recovery” pruning paradigm, with a significant emphasis on recovery. The unique value of this approach is its ability to reconstruct the distorted output of the pruned model, eliminating the need for expensive and time-consuming retraining.  This method, which relies on a sampled calibration dataset, avoids the need for additional complex engineering efforts. Importantly, our research places a paramount focus on the recovery strategy within this context. While prior studies, such as those by \cite{park2023knowledge}, \cite{kwon2022fast}, and \cite{nova2023gradient}, have predominantly concentrated on pruning criteria, we assert that the recovery strategy is the pivot in retraining-free scenarios. As explained in Section~\ref{sec:compatibility_criteria}, our findings show that even pruning metrics not initially designed for this compression paradigm can be highly effective when paired with a suitable recovery strategy. This highlights the crucial role of recovery strategies in improving the practicality and effectiveness of retraining-free pruning methodologies, which is a facet that existing recovery strategies have failed to adequately address.

\begin{figure*}[t]
\centering
    \subfloat[]{
    \centering
    \includegraphics[width=0.43\linewidth]{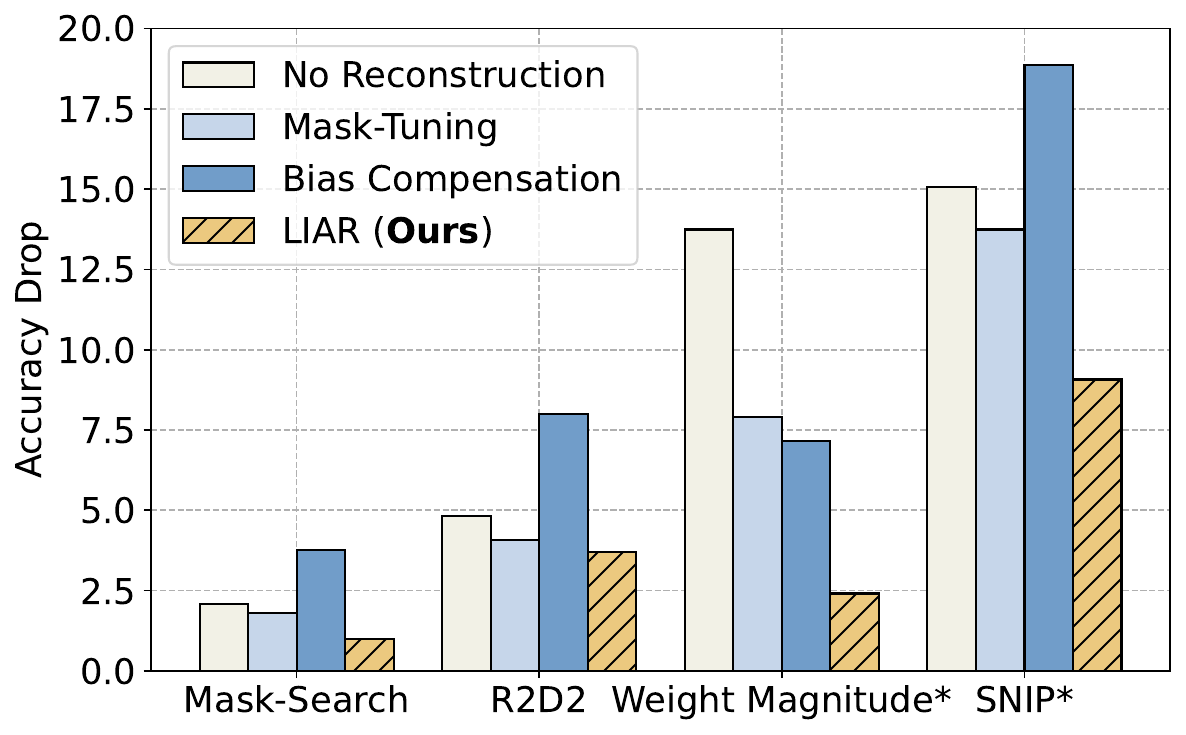}
    \label{fig:intro_result}
    }
    \subfloat[]{
    \centering
    \includegraphics[width=0.55\linewidth]{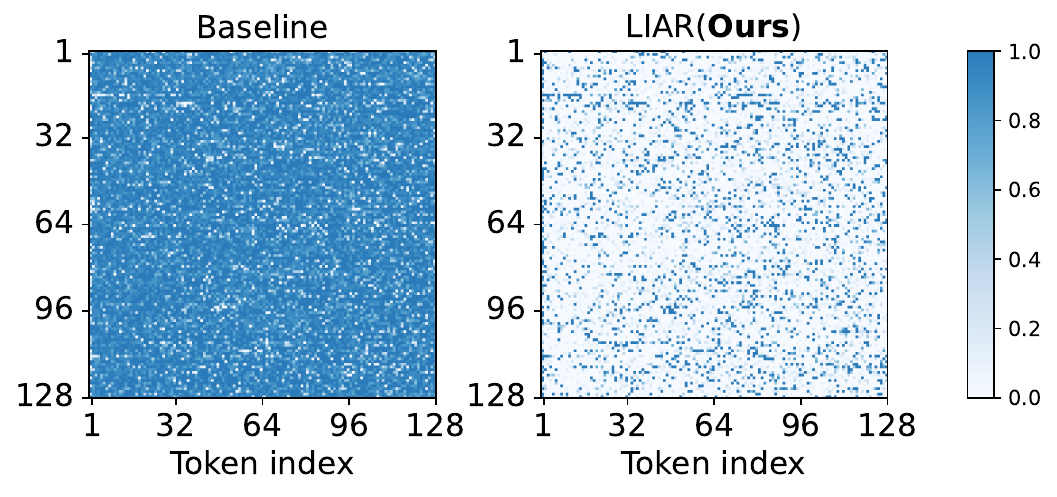}
    \label{fig:single_channel}
    }
    \caption{(a) Accuracy drop on the STS-B task by dropping 70\% FFN neurons of BERT with various pruning criteria (\textit{x-axis}) and reconstruction methods (\textit{legends}). `*' means the retraining-based criteria. (b) Reconstruction error across different tokens and samples by Bias Compensation and LIAR.}
\end{figure*}

However, there is limited research about retraining-free approaches, and most previous works target either encoder-based or decoder-based models exclusively. Additionally, existing retraining-free methods primarily focus on developing better criteria for determining the pruned architecture, with proposed reconstruction techniques often lacking generalizability. As illustrated in Figure~\ref{fig:intro_result}, we applied different algorithms~\cite{kwon2022fast,an2023fluctuation} to reconstruct models pruned using manifold criteria~\cite{kwon2022fast,nova2023gradient,li2017pruning,lee2018snip} and compared the accuracy drop. Our results reveal that existing reconstruction approaches exhibit limited and unstable performance, particularly for retraining-based criteria. Therefore, despite the efficiency of the retraining-free pruning paradigm, its applicability remains significantly restricted.

% \begin{figure}[t]
%     \begin{minipage}[t]{0.53\linewidth}
%     \centering
%     \includegraphics[width=\linewidth]{fig/intro_result.pdf}
%     \vspace{-3mm}
%     \caption{The Performance on the STS-B task by dropping 70\% FFN neurons with various pruning criteria (\textit{x-axis}) and reconstruction approaches (\textit{legends}).}
%     \end{minipage}
%     \hspace{5mm}
%     \begin{minipage}[t]{0.42\linewidth}
%     \centering
%     \includegraphics[width=\linewidth]{fig/stability_linearity_input.pdf}
%     \vspace{-3mm}
%     \caption{The reconstruction error derived by stability-based and linearity-based methods respectively.}
%     \end{minipage}
% \end{figure}

% 总分：stability + linearity, stability is xx, linearity is xx...
To tackle this challenge, we introduce a \textbf{L}inear \textbf{I}nterpolation-based \textbf{A}daptive \textbf{R}ecovery (\textbf{LIAR}) framework, an efficient and effective distortion reconstruction framework for retraining-free structured pruning. In this framework, we reformulate the reconstruction problem as the estimation of the pruned output, and utilize the reserved modules to approximate the pruned ones, which achieves a much lower reconstruction error than the existing work as shown in Figure~\ref{fig:single_channel}. Through this framework, our reconstruction algorithm can not only be applied to both encoder- and decoder-based models, but also generalize better on extensive pruning criteria that are not originally designed for such a retraining-free paradigm, which is exhibited in Figure~\ref{fig:intro_result}. In this way, it largely boosts the application potential for efficient model compression.

% which is based on the to attributes of the uncompressed model: stability and linearity. We observe that the hidden states partly exhibit a stationary pattern, which is called stability. Apart from the stable component, we also find that the extra volatile part of one channel can be estimated by linear fitting with others, and we name this manifestation linearity.
% Based on the above discoveries, we reformulate the recovery problem as the estimation of the pruned output and utilize the stability and linearity to reconstruct the pruned output with the remained ones by linear fitting. Finally, the estimated correlations between the masked and unmasked channels are merged into the preserved weight matrix by interpolation.

% recovery capability -> compression performance
To evaluate the compression performance enhancement capability of our LIAR framework, we conduct experiments on both BERT$_{\rm BASE}$ and LLaMA models family for GLUE, SQuAD, WikiText, 7 common sense reasoning benchmarks for sequence classification, question answering (QA), language modeling and zero-shot performance validation respectively. 
% 4 categorized for performance validation.
We also assess the performance of LIAR for different modules and pruning criteria to investigate the generalizability of our reconstruction framework. Our contributions can be summarized as follows:
\begin{itemize}
    \item \textbf{Framework}. We reformulate the distortion reconstruction problem, and propose LIAR, an efficient, effective, and unsupervised reconstruction framework that requires no backward propagation or retraining. It utilizes the preserved modules to approximate the impacts of the pruned ones, thereby achieving efficient and accurate performance reconstruction.
    \item \textbf{Performance}. We show that LIAR achieves the highest accuracy compared with existing state-of-the-art (SOTA) retraining-free pruning approaches on both BERT$_{\rm BASE}$ and LLaMA family models on 4 categorized benchmarks. Notably, LIAR achieves 98\% accuracy for BERT$_{\rm BASE}$ with 50\% parameters pruned.
    \item \textbf{Generalization}. We conduct extensive experiments to verify that LIAR is capable of generalizing across different modules and criteria. This increases the utility of both retraining-based and retraining-free criteria within the retraining-free paradigm, thereby expanding their range of applications.

\end{itemize}

% \begin{figure}[t]
%     \centering
%     % \includegraphics[width=\linewidth, height=0.2\linewidth]{example-image-duck}
%     \includegraphics[width=0.5\linewidth]{fig/liar_intro.pdf}
%     % \vspace{-3mm}
%     \caption{duck}
%     \label{fig:intro}
% \end{figure}

\section{Related Works}
% 感觉应该更侧重于解决了xx问题：以前的解决方案、存在xx问题、我们如何解决
\label{sec:relat}

\begin{table*}[h]
    \centering
    % \vspace{-2em}
    \caption{Comparison between various reconstruction methods for PLMs concerning different aspects. \textcolor[RGB]{34,139,34}{\Checkmark} and \textcolor{red}{\XSolidBrush} represent whether the method had the specific feature or not.}
    \resizebox{0.9\linewidth}{!}{
    \begin{tabular}{ccccc}
        \toprule
        Reconstruction Method & \makecell[c]{Encoder-based\\Models} & \makecell[c]{Decoder-based\\Models}  & \makecell[c]{High Pruning\\Ratio} & \makecell[c]{Arbitrary Pruning\\Criteria} \\
        \midrule
        No reconstruction & \textcolor{red}{\XSolidBrush} & \textcolor{red}{\XSolidBrush} & \textcolor{red}{\XSolidBrush} & \textcolor{red}{\XSolidBrush} \\
        Mask-Tuning~\cite{kwon2022fast,nova2023gradient} & \textcolor[RGB]{34,139,34}{\Checkmark} & \textcolor{red}{\XSolidBrush} & \textcolor{red}{\XSolidBrush} & \textcolor[RGB]{34,139,34}{\Checkmark} \\
        Bias Compensation~\cite{an2023fluctuation} & \textcolor{red}{\XSolidBrush} & \textcolor[RGB]{34,139,34}{\Checkmark} & \textcolor{red}{\XSolidBrush} & \textcolor{red}{\XSolidBrush} \\
        \midrule
        LIAR (\textbf{Ours}) & \textcolor[RGB]{34,139,34}{\Checkmark} & \textcolor[RGB]{34,139,34}{\Checkmark} & \textcolor[RGB]{34,139,34}{\Checkmark} & \textcolor[RGB]{34,139,34}{\Checkmark} \\
        \bottomrule
    \end{tabular}
    }
    % \vspace{-0.8em}
    
    \label{tab:comparison}
\end{table*}

\subsection{Network Pruning for Language Models}
Network pruning is a widely applicable compression technique, whose key point is to remove the redundant weight or modules from the original network \cite{vaswani2017attention} and reserve the salient ones \cite{lecun1989optimal,hassibi1993optimal}. It is broadly categorized from the granularity aspect into structured and unstructured pruning. Unstructured pruning \cite{gale2019state,sun2023simple,frantar2023sparsegpt} performs at the individual weight level, which brings about larger sparsity but fails to accelerate the model and reduce the storage cost without requiring additional hardware support. By contrast, structured pruning \cite{he2023structured,fan2019reducing,voita2019analyzing} removes a group of weights, such as an entire channel, head, layer, and so on, therefore providing a more hardware-friendly solution, enhancing the lower inference latency and memory demands, so we focus on structured pruning in this paper.

% \subsection{Retraining-free Compression}
%The conventional retraining-based paradigm is to first compress the original model through manifold criteria, and then regain the performance of the lightweight model by retraining \cite{sun2019patient,lagunas2021block,han2015deep,kurtic2022optimal}. Whereas, as the model scale emerges rapidly with the significant achievement of LLMs \cite{brown2020language,kaplan2020scaling,zhang2022opt}, such a traditional pipeline may be too expensive to be practical, which motivates the exploration of retraining-free compression approaches. The existing works in this genre primarily focus on quantization \cite{dettmers2022llm,frantar2022gptq,li2021brecq}, which has also recently been successfully extended to pruning~\cite{kwon2022fast,nova2023gradient,an2023fluctuation}.

% with the proposed research about retraining-free pruning~\cite{kwon2022fast,nova2023gradient} mostly designed for smaller encoder-based models like BERT \cite{devlin2018bert}. Thus these methods are probably infeasible for the SOTA encoder-decoder or decoder-based LLMs, which are more sensitive to the parameter change. A previous study proposed FLAP \cite{an2023fluctuation} to address this issue, but it exhibits fairly unstable performance for the encoder-based model BERT. To address these issues, we propose a powerful reconstruction framework, which enhances the existing retraining-free pruning algorithms and is universally applicable to various model structures as Table~\ref{tab:comparison} shows. 
The conventional retraining-based paradigm involves compressing the original model using various criteria followed by retraining to restore performance \cite{sun2019patient,lagunas2021block,han2015deep,kurtic2022optimal}. However, as the size and complexity of LLMs rapidly increase \cite{brown2020language,kaplan2020scaling,zhang2022opt}, this conventional approach becomes impractical and costly, prompting the need for retraining-free compression techniques. Recent developments in this area have primarily centered around quantization \cite{dettmers2022llm,frantar2022gptq,li2021brecq} and have expanded to include pruning methods \cite{kwon2022fast,nova2023gradient,an2023fluctuation} that eliminate the need for retraining. In this paper, our work targets enhancing the performance of the retraining-free pruning paradigm, which can reduce the model size, lower the memory consumption, accelerate the inference, and be orthogonal and compatible with quantization for further compression simultaneously.

\subsection{Distortion Reconstruction for Retraining-free Pruning}
% distortion recovery vs. performance recovery
%For network pruning, the framework of retraining-free methods \cite{park2023knowledge} is to replace the retraining stage with recovering the output distortion to retain the model capability as much as possible. 
% However, there are limited related works compared to retraining-based pruning, and most of previous works focus more on the pruning criteria than on distortion reconstruction.
%\cite{kwon2022fast} proposed Mask-Tuning as the reconstruction technique by rescaling the mask, which is also utilized by KCM \cite{nova2023gradient}. However, it only evaluates the feasibility of encoder-based models and fails to retain the performance under high pruning ratios. FLAP \cite{an2023fluctuation} proposed the bias compensation to restore the distorted layer output, but it is specifically designed for its pruning metric, and may not universally generalize to extensive retraining-based effective criteria and even exhibits unstable performance as explained in \ref{fig:intro_result}. Our reconstruction framework successfully addresses these issues. It not only significantly reduces the performance gap between the original and pruned models, particularly at high pruning ratios, but holds stable generality to various pruning criteria as summarized in Table~\ref{tab:comparison}.
In the context of network pruning, retraining-free approaches such as those proposed by \cite{park2023knowledge} seek to mitigate output distortion instead of retraining to maintain as much of the model's original capability as possible. Mask-Tuning, introduced by \cite{kwon2022fast} and adopted by KCM \cite{nova2023gradient}, involves rescaling the mask as a reconstruction technique. While it tests the limits of encoder-based models, it struggles to maintain performance at high pruning ratios. FLAP \cite{an2023fluctuation} introduced a bias compensation method to correct the distorted output of pruned layers. However, it is tailored to its specific pruning metric and may not broadly apply to other effective pruning criteria, sometimes leading to unstable performance as discussed in Figure~\ref{fig:intro_result}. Our proposed framework overcomes these limitations by significantly narrowing the performance gap between the original and pruned models, especially at higher pruning ratios, and maintains stable effectiveness across various pruning standards as detailed in Table~\ref{tab:comparison}.

% \begin{figure}[t]
%     \centering
%     % \includegraphics[width=\linewidth, height=0.3\linewidth]{example-image-duck}
%     \includegraphics[width=\linewidth]{fig/single_channel.pdf}
%     % \vspace{-3mm}
%     \caption{Reconstruction error across different tokens and samples by stability- and linearity-based approaches.}
%     \label{fig:single_channel}
% \end{figure}

\section{Preliminaries}
\paragraph{Post-training pruning.}
Given a computational constraint and a sampled calibration dataset, the target of post-training pruning is to prune a well-optimized model to satisfy the constraint. Considering a fine-tuned model $\mathcal{M}$ and a constraint $\mathcal{C}$, optimal structured pruning is usually defined with respect to minimizing the accuracy loss:
\begin{equation}
    \mathop{\arg\min}\limits_{\mathcal{M}} \mathcal{L}(\mathcal{M}) \quad\quad s.t. \quad Cost(\mathcal{M}) \leq \mathcal{C},
\end{equation}
where $Cost(.)$ is a calculation function of computational complexity. In the retraining-free context, $\mathcal{L}(\mathcal{M})$ can also be replaced with feature map loss or other metrics~\cite{an2023fluctuation,nova2023gradient}. 

\paragraph{Layer-wise Pruning.}
For post-training pruning, globally solving the pruning problem is challenging and technically infeasible due to the enormous neural architecture search expense. A practical solution for it is to split the full-model problem into layer-wise subproblems as \cite{frantar2023sparsegpt} demonstrated. Given an input $\mathbf{X}^\ell$ of shape $(N, T, C_{in})$ with $N$ instances and sequence length $T$ for layer $\ell$, a weight $\mathbf{W}^\ell$ of shape $(C_{in}, C_{out})$ and bias $\mathbf{B}^\ell$ of shape $(C_{out,})$, the quality of the layer-wise solution for structured pruning is usually measured by the $\ell_2$-error between the original output  and the pruned output as 
\begin{equation}
    % \resizebox{.91\hsize}{!}{
    \small
    \mathbf{M}^\ell, \widehat{\mathbf{W}}^\ell = \mathop{\arg\min}\limits_{\mathbf{M}^{\ell},\widehat{\mathbf{W}}^{\ell}} \left\|\mathbf{X}^{\ell} \mathbf{W}^{\ell} + \mathbf{B}^\ell - \mathbf{X}^{\ell}\left(\mathbf{M}^{\ell} \odot \widehat{\mathbf{W}}^{\ell}\right) - \widehat{\mathbf{B}}^\ell \right\|_2^2,
    % }
    \label{eq:WX-MWX}
\end{equation}
where $\mathbf{M}^\ell\in\mathbb{R}^{C_{in}}$ denotes the mask vector of layer $\ell$, $\|.\|^2_2$ denotes the $\ell_2$-error, $\widehat{\mathbf{W}}^\ell$ and $\widehat{\mathbf{B}}^\ell$ are the weight and bias of the pruned model and probably updated from $\mathbf{W}^\ell$ and $\mathbf{B}^\ell$ through retraining.

% In this section, we will first reformulate the distortion recovery problem, then introduce the significant attributes utilized in our methodology, and finally explain and formulate the proposed LIAR framework, which is shown in Figure~\ref{fig:overview}.

% \paragraph{Reconstruction Problem Reformulation}

\begin{figure*}[t]
    \centering
    \includegraphics[width=\linewidth]{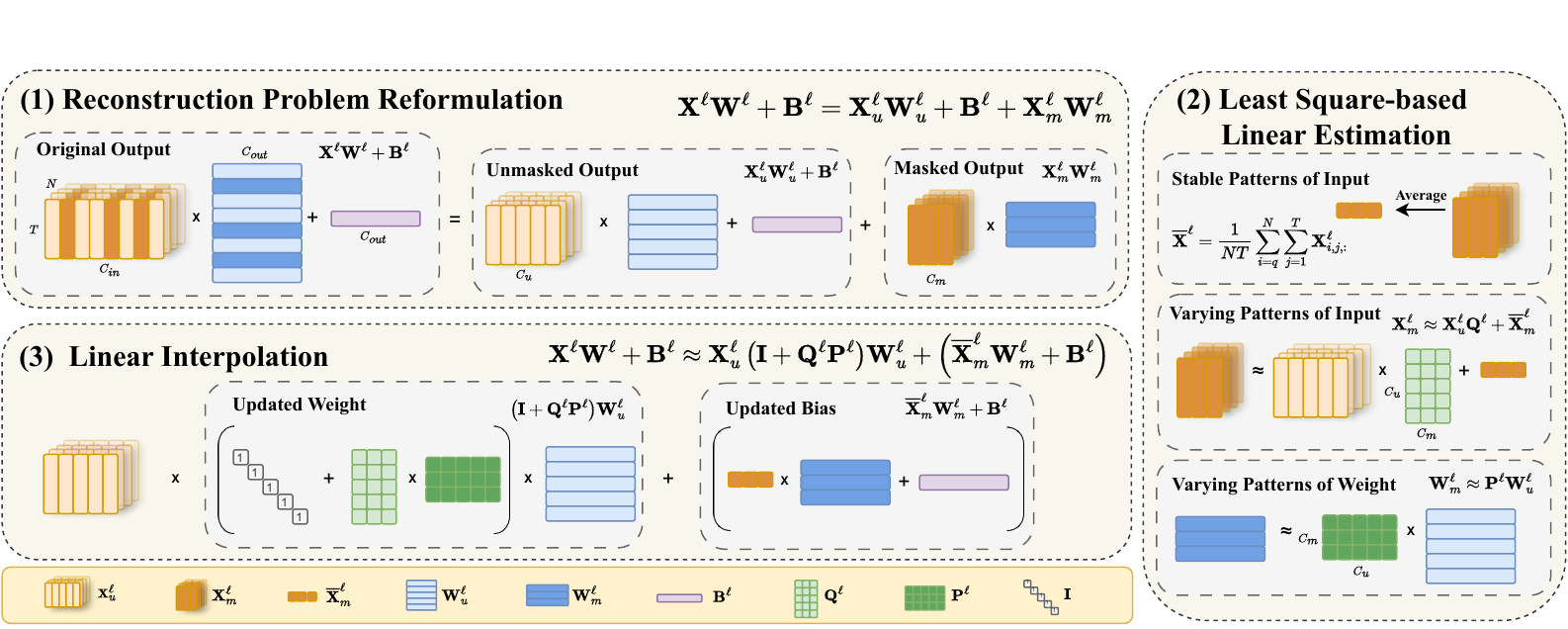}
    \vspace{-3mm}
    \caption{Overview of LIAR framework. The original output is first reformulated by \textbf{(1) Reconstruction Problem Reformulation}, and then the masked output is estimated by \textbf{(2) Least Square-based Linear Estimation}. Finally, the model weight and bias are updated with \textbf{(3) Linear Interpolation}.}
    \label{fig:overview}
\end{figure*}

\section{LIAR: Linear Interpolation-based Adaptive Recover}
\label{sec:liar}

\subsection{Motivation}

To conduct a more detailed analysis of the reconstruction problem, we would like to first reformulate the pruning and reconstruction process. 
% In the context of retraining-free pruning, the updated $\widehat{\mathbf{W}}^\ell$ is not obtained by retraining but by reconstruction. Therefore, 
Considering a transformation function with weight $\mathbf{W}^\ell$ and bias $\mathbf{B}^\ell$, we reformulate the original output $\mathbf{X}^\ell \mathbf{W}^\ell + \mathbf{B}^\ell$ to investigate how the counterparts derive the output and result in the distortion after pruning:
\begin{equation}
    \small
    \begin{matrix}
    \mathbf{X}^\ell \mathbf{W}^\ell + \mathbf{B}^{\ell} =
     \underbrace{\mathbf{X}_u^\ell \mathbf{W}_u^\ell + \mathbf{B}^\ell}_{\text{unmasked output}} 
     + \underbrace{\mathbf{X}_m^\ell \mathbf{W}_m^\ell}_{\text{masked output}},
     \end{matrix}
     \label{eq:XW=XuWx+B+XmWm}
\end{equation}
where $\mathbf{X}_u^\ell$ of shape $(N,T,C_u)$ and $\mathbf{W}_u^\ell$ of shape $(C_u,C_{out})$ denote the unmasked input activation and weight. By contrast, the product of the masked input $\mathbf{X}_m^\ell$ of shape $(N, T, C_m)$ and weight $\mathbf{W}_m^\ell$ of shape $(C_m, C_{out})$, leads to the output distortion and performance degradation if pruned directly, and it is also the retraining or reconstruction target to guarantee the output fidelity.

In the circumstance of the conventional pruning, the masked output is regained by retraining the unmasked weight $\mathbf{W}_u^\ell$ and bias $\mathbf{B}^\ell$, while we target resolving this problem by approximating the masked component with the unmasked output. The reconstruction error of $k$-th channel of layer $\ell$ is defined as 
\begin{equation}
    \varepsilon^\ell_k = \frac{\Vert \widehat{\mathbf{X}}^\ell_{:,:,k} - \mathbf{X}^\ell_{:,:,k} \Vert^2_2}{\Vert \mathbf{X}^\ell_{:,:,k} \Vert^2_2},
\end{equation}
 and a lower reconstruction error typically indicates better performance retention.

Previous work~\cite{an2023fluctuation} approximated the pruned output by utilizing the stability of part of the channels, which exhibit fairly stable patterns given different samples and thus can be compensated with a bias term. The reconstructed representation for channel $k$ is the averaged value, which is formulated as $\overline{\mathbf{X}}^{\ell}_{k}=\frac{1}{NT} \sum\limits_{i=1}^{N} \sum\limits_{j=1}^{T} \mathbf{X}^\ell_{i,j,k}$. Such a stability-based approach may work for some stable channels, but fails to reconstruct the ones with high fluctuations as demonstrated in Figure \ref{fig:single_channel} (left part).

\subsection{Method}

To address the issue of instability in certain channels, we propose to approximate the varying patterns that cannot be compensated by the averaged value. Specifically, we reconstruct the varying patterns of pruned channel $k$ with a linear combination of the others, which is formulated as $\widehat{\mathbf{X}}^\ell_{:,:,k} = \overline{\mathbf{X}}^{\ell}_{k} + \sum\limits_{\substack{l=1, l\neq k}}^{C_{in}} m_{l} \mathbf{X}_{:,:,l}$,
% \begin{equation}
%     \widehat{\mathbf{X}}^\ell_{:,:,k} = \overline{\mathbf{X}}^{\ell}_{k} + \sum\limits_{\substack{l=1, l\neq k}}^{C_{in}} m_{l} \mathbf{X}_{:,:,l},
% \end{equation}
where $m_l\in\mathbb{R}$ is a scalar that determines the contribution of channel $l$ to reconstruct the channel $k$. It is calculated by the least square algorithm. As shown in Figure \ref{fig:single_channel} (right part), compared to the stability-based one, our method achieves much lower reconstruction error.

% ulteriorly
Furthermore, we introduce our \textbf{L}inear \textbf{I}nterpolation-based \textbf{A}daptive \textbf{R}econstruction (\textbf{LIAR}) framework. This method approximates the pruned $\mathbf{X}^\ell_m$ and $\mathbf{W}^\ell_m$ using the remaining $\mathbf{X}^\ell_u$ and $\mathbf{W}^\ell_u$. We compute the transformation matrices and apply linear interpolation to the preserved weight matrix $\mathbf{W}^\ell_u$ to effectively reconstruct the distortion introduced by pruning.

To be specific, we first define a transformation matrix $\mathbf{Q}^\ell$ to approximate the varying patterns of $\mathbf{X}^\ell_m$ by $\mathbf{X}^\ell_u$ as
\begin{equation}
\small
    \mathbf{X}^\ell_m \approx \mathbf{X}^\ell_u \mathbf{Q}^\ell + \overline{\mathbf{X}}^\ell_m,
    \label{eq:Xm=XuQ+Xm}
\end{equation}
where $\mathbf{Q}^\ell \in \mathbb{R}^{C_u \times C_m}$ is derived by the least square algorithm to minimize the reconstruction error: 
\begin{equation}
\small
    \mathbf{Q}^\ell = \mathop{\arg\min}\limits_{\mathbf{\mathbf{Q}^\ell}}  \left\Vert \mathbf{X}^\ell_m - \left( \mathbf{X}^\ell_u \mathbf{Q}^\ell + \overline{\mathbf{X}}^\ell_m \right) \right\Vert^2_2.
    \label{eq:argmin_Xm-XuQ-Xm}
\end{equation}

%A similar procedure is easily adapted to recover the masked weight matrix $\mathbf{W}^\ell_m$ as
%\begin{equation}
%    \mathbf{W}^\ell_m \approx \mathbf{P}^\ell \mathbf{W}^\ell_u,
%    \label{eq:Wm=PWu+Wm}
%\end{equation}
%where $\mathbf{P}^\ell \in \mathbb{R}^{C_m \times C_u}$, which is resolved as Equation (\ref{eq:argmin_Xm-XuQ-Xm}) similarly. We do not add a bias term as Equation (\ref{eq:Xm=XuQ+Xm}) because the stability is not observed for weight matrices, and we have tried to add it but did not observe significant improvement.
The masked weight matrix $\mathbf{W}^\ell_m$ can be reconstructed following a similar procedure:
\begin{equation}
\small
    \mathbf{W}^\ell_m \approx \mathbf{P}^\ell \mathbf{W}^\ell_u,
    \label{eq:Wm=PWu+Wm}
\end{equation}
where $\mathbf{P}^\ell \in \mathbb{R}^{C_m \times C_u}$ is determined in the same way as Equation~(\ref{eq:argmin_Xm-XuQ-Xm}). We do not include a bias term as in Equation~(\ref{eq:Xm=XuQ+Xm}) because stability was not observed for the weight matrices.

% \subsection{Linear Interpolation-based Adaptive Recovery (LIAR) Framework}
Having successfully reconstructed $\mathbf{X}^\ell_m$ and $\mathbf{W}^\ell_m$, 
% by Equations~(\ref{eq:Xm=XuQ+Xm}) and (\ref{eq:Wm=PWu+Wm}) respectively
we can now seamlessly combine them to achieve a distortion-free output. Consequently, $\mathbf{X}^\ell_m \mathbf{W}^\ell_m$ is redefined as:
\begin{equation}
\small
\begin{aligned}
    \mathbf{X}^\ell_m \mathbf{W}^\ell_m 
    &\approx (\mathbf{X}^\ell_u \mathbf{Q}^\ell + \overline{\mathbf{X}}^\ell_m) \mathbf{W}^\ell_m \\
    &= \mathbf{X}^\ell_u \mathbf{Q}^\ell \mathbf{W}^\ell_m + \overline{\mathbf{X}}^\ell_m \mathbf{W}^\ell_m \\
    % &\approx \mathbf{X}^\ell_u \mathbf{Q}^\ell (\mathbf{P}^\ell \mathbf{W}^\ell_u  + \overline{\mathbf{W}}^\ell_m^\top)  + \overline{\mathbf{X}}^\ell_m \mathbf{W}^\ell_m \\
    &= \mathbf{X}^\ell_u \mathbf{Q}^\ell \mathbf{P}^\ell \mathbf{W}^\ell_u + \overline{\mathbf{X}}^\ell_m \mathbf{W}^\ell_m.
    \label{eq:XmWm=xx}
\end{aligned}
\end{equation}

% \begin{table*}[htbp]
%     \centering
%     \caption{Perplexity of the pruned LLaMA family models on WikiText. `Bias Comp.' represents Bias Compensation reconstruction.}
%     \resizebox{0.6\linewidth}{!}{
%         \begin{tabular}{l | l | l | c c c  }
%             \toprule
%             \makecell[c]{Pruning\\ratio} & Criteria & Recovery & 7B & 13B & 30B \\
%             \midrule
%             Dense & - & - & 12.62 & 10.81 & 9.11 \\
%             \midrule
%             \multirow{6}{*}{20\%} & \multirow{3}{*}{LLM-Pruner} & Naive & - & 16.59 & 12.35 \\
%             & & Bias Comp. & 18.15 & 15.92 & 11.97 \\
%             & & LIAR & \textbf{15.92} & \textbf{13.75} & \textbf{11.67} \\
%             \cmidrule(lr){2-6}
%             & \multirow{3}{*}{FLAP} & Naive & 16.15 & 14.75 & 11.96 \\
%             & & Bias Comp. & 14.62 & 14.17 & 11.46 \\
%             & & LIAR & \textbf{14.07} & \textbf{12.71} & \textbf{10.93} \\
%             \midrule
%             \multirow{6}{*}{50\%} & \multirow{3}{*}{LLM-Pruner} & Naive & 112.44 & 76.40 & 36.16 \\
%             & & Bias Comp. & 82.60 & 44.82 & 26.66 \\
%             & & LIAR & \textbf{43.96} & \textbf{30.15} & \textbf{19.58} \\
%             \cmidrule(lr){2-6}
%             & \multirow{3}{*}{FLAP} & Naive & 52.74 & 36.37 & 26.11 \\
%             & & Bias Comp. & 31.80 & 24.83 & 20.54 \\
%             & & LIAR & \textbf{25.43} & \textbf{21.11} & - \\
%             \bottomrule
%         \end{tabular}
%     }
%     \label{tab:exp_ppl}
% \end{table*}

Then we substitute Equation~(\ref{eq:XmWm=xx}) to (\ref{eq:XW=XuWx+B+XmWm}) and obtain
\begin{equation}
\small
% \resizebox{.89\hsize}{!}{
    \begin{aligned}
    \mathbf{X}^\ell \mathbf{W}^\ell + \mathbf{B}^\ell 
    & \approx \mathbf{X}^\ell_u \mathbf{W}^\ell_u + \mathbf{B}^\ell + \mathbf{X}^\ell_u \mathbf{Q}^\ell \mathbf{P}^\ell \mathbf{W}^\ell_u + \overline{\mathbf{X}}^\ell_m \mathbf{W}^\ell_m \\
    &= \mathbf{X}^\ell_u \left(\mathbf{I} + \mathbf{Q}^\ell \mathbf{P}^\ell \right) \mathbf{W}^\ell_u + (\overline{\mathbf{X}}^\ell_m \mathbf{W}^\ell_m + \mathbf{B}^\ell),
    \end{aligned}
% }
\label{eq:final_equation}
\end{equation}
where $\mathbf{I}$ of shape $(C_u, C_u)$ is the identity matrix, the preserved weight $\mathbf{W}_u^\ell$ and bias $\mathbf{B}^\ell$ are updated to $(\mathbf{I} + \mathbf{Q}^\ell \mathbf{P}^\ell)\mathbf{W}^\ell_u$ and $(\overline{\mathbf{X}}^\ell_m \mathbf{W}^\ell_m + \mathbf{B}^\ell)$, respectively. By Interpolating the transformation matrices to the preserved weights and bias, we achieve to reconstruct the output distorted by direct pruning. 

\subsection{Framework}
We visualize our framework in Figure~\ref{fig:overview}, which includes (1) Reconstruction Problem Reformulation, (2) Least-Square-based Linear Estimation, and (3) Linear Interpolation. Specifically, considering a function with weight $\mathbf{W}^\ell$ and bias $\mathbf{B}^\ell$, we first reformulate the original output and split it into the masked and unmasked ones, denoted as $\mathbf{X}^\ell_m \mathbf{W}^\ell_m$ and $\mathbf{X}^\ell_u \mathbf{W}^\ell_u + \mathbf{B}^\ell$. Secondly, we estimate the stable and varying patterns of the masked input $\mathbf{X}^\ell_m$ and weight $\mathbf{W}^\ell_m$ and obtain the transformation matrices $\mathbf{Q}^\ell$ and $\mathbf{P}^\ell$ by the least square algorithm. Finally, with the obtained transformation matrices, we apply linear interpolation to the preserved weight and bias to reconstruct the pruned output. 
The comprehensive details of our method are further described in Appendix~\ref{apx:algorithm}.

\section{Experiments}
\label{sec:exp}

\subsection{Experimental Setup}
\label{sec:exp_setup}

\paragraph{Models.}
We conduct experiments on two representative and widely used models with different architectures and sizes: BERT$_{\rm BASE}$~\cite{devlin2018bert} (0.1B) and LLaMA family models~\cite{touvron2023llama} (7B, 13B and 30B), which are encoder- and decoder-based respectively.

\paragraph{Tasks.}
To verify the widespread adoption of our proposed framework, we evaluate LIAR on four categories of capabilities: the sequence classification performance on the GLUE benchmark \cite{wang2018glue}, the question-answering capability on SQuAD$_{1.1}$ \cite{rajpurkar2016squad} and SQuAD$_{2.0}$ \cite{rajpurkar2018know} datasets, the language modeling capability on the WikiText-2 \cite{merity2016pointer} dataset, and the zero-shot performance across seven common sense reasoning tasks~\cite{clark2019boolq,bisk2020piqa,zellers2019hellaswag,sakaguchi2021winogrande,clark2018think,mihaylov2018can}.

\paragraph{Baselines.}
To validate the effectiveness of our reconstruction algorithm, we compare the performance reconstructed with various state-of-the-art structured pruning approaches. These include retraining-based method like LLM-Pruner \cite{ma2023llm}, and retraining-free algorithms, such as Mask-Tuning~\cite{kwon2022fast}, KCM~\cite{nova2023gradient}, and FLAP~\cite{an2023fluctuation}. The baseline reconstruction methods include Mask-Tuning~\cite{kwon2022fast} and Bias Compensation~\cite{an2023fluctuation}.

\paragraph{Pruning Criteria \& Reconstruction.}
Finally, to evaluate the generalization ability to various pruning criteria across different pruning ratios, we test LIAR with two representative retraining-based pruning metrics: Weight Magnitude~\cite{li2017pruning} and SNIP~\cite{lee2018snip}. These metrics use the magnitude and the first derivative of the model weight respectively, and are known as zero- and first-order criteria.

\paragraph{Implementation Details.}
We implement the framework using the Pytorch
% \footnote{\url{https://github.com/pytorch}}
\cite{paszke2019pytorch} and Huggingface Transformers library
% \footnote{\url{https://github.com/huggingface/transformers}} 
\cite{wolf2020transformers}. Specifically, the least square algorithm is implemented by the linear solver\footnote{torch.linalg.lstsq} to derive the solution for Equation~(\ref{eq:argmin_Xm-XuQ-Xm}). All the experiments for both the BERT$_{\rm BASE}$ and LLaMA models are conducted on one single NVIDIA Tesla A100 80G GPU. Please see Appendix~\ref{apx:experimental_setup} and \ref{apx:detailed_implementation} for more experimental settings and implementation details.

\begin{table*}[t]
    \centering
    \caption{Sequence classification and QA performance of the pruned BERT$_{\rm BASE}$ on the GLUE benchmark and SQuAD datasets. ``Naive'' represents the results without retraining and reconstruction.} 
    \resizebox{\linewidth}{!}{
        \begin{tabular}{l | l | l | c c c c c c c c | c}
            \toprule
            Ratio & Criterion & Reconstruction & MNLI$\uparrow$ & MRPC$\uparrow$ & STS-B$\uparrow$ & SST-2$\uparrow$ & QNLI$\uparrow$ & QQP$\uparrow$ & SQuAD$_{1.1}$$\uparrow$ & SQuAD$_{2.0}$$\uparrow$ & Average$\uparrow$ \\
            \midrule
            Dense & - & - & 84.53 & 86.27 & 88.59 & 93.12 & 91.41 & 91.00 & 88.48 & 76.82 & 87.53 \\
            \midrule
            \multirow{6}{*}{30\%} & \multirow{3}{*}{KCM} & Naive & 75.87 & 66.91 & 86.65 & 91.28 & 84.53 & 87.23 & 75.62 & 38.70 & 75.85 \\
             & & KCM & 80.49 & \textbf{85.54} & 86.51 & 92.09 & 88.12 & 89.41 & 84.62 & 72.66 & 84.93 \\
            & & LIAR & \textbf{82.90} & 84.31 & \textbf{87.54} & \textbf{92.20} & \textbf{89.09} & \textbf{90.16} & \textbf{86.13} & \textbf{72.90} & \textbf{85.65}\\
            \cmidrule(lr){2-12}
            & \multirow{3}{*}{Mask-Tuning} & Naive & 83.63 & 84.31 & 88.34 & 92.78 & 90.23 & 90.75 & 87.08 & 75.49 & 86.58 \\
            & & Mask-Tuning & 83.05 & \textbf{86.76} & 88.38 & 92.66 & 90.79 & 90.72 & 87.55 & \textbf{76.06} & 87.00 \\
            & & LIAR & \textbf{84.10} & 86.52 & \textbf{88.63} & \textbf{93.00} & \textbf{91.01} & \textbf{90.85} & \textbf{87.91} & 75.56 & \textbf{87.20} \\

            \midrule
            \multirow{6}{*}{50\%} & \multirow{3}{*}{KCM} & Naive & 50.20 & 34.31 & 82.63 & 64.11 & 79.33 & 73.59 & 47.63 & 24.23 & 57.00 \\
            & & KCM & 50.20 & 34.31 & 82.63 & 64.11 & 79.33 & 73.59 & 48.99 & 24.23 & 57.17\\
            & & LIAR & \textbf{76.36} & \textbf{78.92} & \textbf{83.65} & \textbf{90.25} & \textbf{80.91} & \textbf{87.98} & \textbf{76.36} & \textbf{53.10} & \textbf{78.44} \\
            \cmidrule(lr){2-12}
            & \multirow{3}{*}{Mask-Tuning} & Naive & 76.67 & 81.37 & 86.81 & 89.33 & 87.00 & 88.69 & 77.18 & 56.70 & 80.47 \\
            & & Mask-Tuning & 80.87 & 83.58 & 86.92 & 91.74 & \textbf{88.94} & 89.53 & 81.67 & 68.26 & 83.94 \\
            & & LIAR & \textbf{82.87} & \textbf{86.27} & \textbf{87.92} & \textbf{92.43} & 88.76 & \textbf{90.17} & \textbf{85.45} & \textbf{71.81} & \textbf{85.71} \\
            \bottomrule
        \end{tabular}
    }
    \label{tab:exp_bert}
\end{table*}

\subsection{Classification \& QA Tasks Performance}

We first establish the classification and QA performance on the GLUE and SQuAD benchmarks for the encoder-based BERT$_{\rm BASE}$ model with SOTA retraining-free methods: Mask-Tuning~\cite{kwon2022fast} and KCM~\cite{nova2023gradient}. To be specific, we only adopt their pruning criteria and compare the performance recoverd by LIAR with naive pruning (no reconstruction) and baselines.

%As shown in Table~\ref{tab:exp_bert}, the performance with LIAR significantly and consistently outperforms both baselines. Notably, it retains 99.6\% accuracy by removing 20\% encoder parameters, and prunes 50\% parameters within 2\% performance degradation compared to the uncompressed BERT$_{\rm BASE}$ without any retraining. Furthermore, LIAR is able to improve the performance derived by naive pruning without recovery immensely by, for example, 37.6\% for KCM when p=50\%.
As shown in Table~\ref{tab:exp_bert}, the performance with LIAR significantly and consistently outperforms both baselines. Notably, it retains 99.6\% accuracy even after removing 20\% of encoder parameters. It also prunes 50\% of parameters with only a 2\% performance degradation compared to the uncompressed BERT$_{\rm BASE}$, without any retraining. Furthermore, LIAR greatly enhances the performance derived by naive pruning without recovery. For instance, it improves performance by 37.6\% for KCM at a 50\% pruning ratio.

\begin{table*}[t]
    \centering
    \caption{Perplexity of the pruned LLaMA-7B, LLaMA-13B and LLaMA-30B on the WikiText.}
    \resizebox{\linewidth}{!}{
        \begin{tabular}{c | l | l | c c c || c | l | l | c c c }
            \toprule
            Ratio & Criterion & Reconstruction & 7B$\downarrow$ & 13B$\downarrow$ & 30B$\downarrow$ & Pruning ratio & Criterion & Reconstruction & 7B$\downarrow$ & 13B$\downarrow$ & 30B$\downarrow$ \\
            \midrule
            Dense & - & - & 12.62 & 10.81 & 9.11 & Dense & - & - & 12.62 & 10.81 & 9.11\\
            \midrule
            \multirow{6}{*}{20\%} & \multirow{3}{*}{LLM-Pruner} & Naive & 19.28 & 16.59 & 12.35 & \multirow{6}{*}{50\%} & \multirow{3}{*}{LLM-Pruner} & Naive & 112.44 & 76.40 & 36.16 \\
            & & FLAP & 18.15 & 15.92 & 11.97 & & & FLAP & 82.60 & 44.82 & 26.66 \\
            & & LIAR & \textbf{15.92} & \textbf{13.75} & \textbf{11.67} & & & LIAR & \textbf{43.96} & \textbf{30.15} & \textbf{19.58} \\
            \cmidrule(lr){2-6}
            \cmidrule(lr){8-12}
            & \multirow{3}{*}{FLAP} & Naive & 16.15 & 14.75 & 11.96 & & \multirow{3}{*}{FLAP} & Naive & 52.74 & 36.37 & 26.11\\
            & & FLAP & 14.62 & 14.17 & 11.46 & & & FLAP & 31.80 & 24.83 & 20.54  \\
            & & LIAR & \textbf{14.07} & \textbf{12.71} & \textbf{10.93} & & & LIAR & \textbf{25.43} & \textbf{21.11} & \textbf{16.93} \\
            \bottomrule
        \end{tabular}
    }
    \label{tab:exp_ppl}
\end{table*}

\begin{table*}[t]
    \centering
    \caption{Zero-shot task accuracy of the pruned LLaMA-7B on common sense reasoning benchmarks.}
    \resizebox{\linewidth}{!}{
        \begin{tabular}{l | l | l | c c c c c c c | c}
            \toprule
            ratio & Criterion & Reconstruction & ARC-c$\uparrow$ & ARC-e$\uparrow$ & BoolQ$\uparrow$ & HellaSwag$\uparrow$ & OBQA$\uparrow$ & PIQA$\uparrow$ & WinoGrande$\uparrow$ & Average$\uparrow$ \\
            \midrule
            Dense & - & - & 44.71 & 72.85 & 75.02 & 76.20 & 44.40 & 79.16 & 70.01 & 66.05 \\
            \midrule
            \multirow{6}{*}{20\%} & \multirow{3}{*}{LLM-Pruner} & Naive & 38.14 & 63.30 & 57.58 & \textbf{69.13} & 39.80 & 75.35 & 63.77 & 58.15 \\
            & & FLAP & 36.77 & \textbf{65.53} & 61.93 & 68.53 & 41.40 & 75.73 & 63.61 & 59.07 \\
            & & LIAR & \textbf{39.08} & 65.24 & \textbf{66.64} & 67.62 & \textbf{41.80} & \textbf{76.61} & \textbf{64.64} & \textbf{60.23} \\
            \cmidrule(lr){2-11}
            & \multirow{3}{*}{FLAP} & Naive & 38.05 & 64.18 & 52.63 & 71.12 & 40.60 & 77.04 & \textbf{68.82} & 58.92 \\
            & & FLAP & 38.05 & 65.32 & 56.94 & \textbf{71.18} & 39.40 & 76.82 & 68.27 & 59.43 \\
            & & LIAR & \textbf{40.61} & \textbf{68.52} & \textbf{74.37} & 70.00 & \textbf{42.40} & \textbf{77.69} & 67.01 & \textbf{62.94} \\
            \midrule
            \multirow{6}{*}{50\%} & \multirow{3}{*}{LLM-Pruner} & Naive & 25.77 & 34.81 & 46.94 & 36.14 & 30.20 & 60.55 & \textbf{53.12} & 41.08 \\
            & & FLAP & \textbf{26.88} & 36.49 & 42.75 & 37.64 & \textbf{32.00} & 62.62 & 52.49 & 41.55 \\
            & & LIAR & 26.79 & \textbf{43.31} & \textbf{56.73} & \textbf{39.56} & 31.60 & \textbf{63.98} & 52.17 & \textbf{44.88} \\
            \cmidrule(lr){2-11}
            & \multirow{3}{*}{FLAP} & Naive & 29.95 & 45.83 & 59.48 & \textbf{52.96} & \textbf{36.60} & \textbf{67.90} & 57.93 & 50.09 \\
            & & FLAP. & 29.27 & 47.22 & 59.63 & 52.13 & 34.60 & 67.52 & 57.06 & 49.63 \\
            & & LIAR & \textbf{33.96} & \textbf{55.89} & \textbf{62.51} & 50.51 & 35.60 & 67.68 & \textbf{60.22} & \textbf{52.34} \\

            \bottomrule
        \end{tabular}
    }
    
    \label{tab:exp_zero}
\end{table*}

\subsection{Language Modeling \& Zero-shot Tasks Performance}
%Furthermore, to examine the generalizability of LIAR, We also conduct experiments on the decoder-based generative model, LLaMA-7B. Specifically, we apply LIAR on the SOTA retraining-free method for LLMs, FLAP. We compare the accuracy against FLAP, LLM-Pruner, and LLM-Pruner without LoRA finetuning.

%As Table \ref{tab:exp_llama} explains, LIAR serves as a significant recovery framework to retain the performance of the pruned models on the majority of tasks and pruning ratios. Notably, we achieved the best language modeling and zero-shot performance with LIAR even compared to the retraining-based approach. More results under various pruning ratios are shown in Appendix \ref{apx:additional_results_perplexity}.
Moreover, to assess the general applicability of LIAR, we conduct tests on the decoder-based LLaMA family models. Specifically, we apply LIAR to the SOTA approaches for LLMs: LLM-Pruner~\cite{ma2023llm} and FLAP~\cite{an2023fluctuation}. It is noted that as LLM-Pruner is retraining-based, we here adopt the reconstruction method of FLAP to recover its performance as a baseline.

Table~\ref{tab:exp_ppl} demonstrates that LIAR acts as a significant reconstruction framework on the language modeling task, especially at a high pruning ratio. For example, compared to naive pruning, LIAR enhances the performance by 2.56$\times$ and 2.07$\times$ for the LLaMA-7B model with 50\% parameters pruned by LLM-Pruner and FLAP respectively. For zero-shot common sense reasoning tasks, LIAR also enhances the performance of the pruned models on most tasks as shown in Table~\ref{tab:exp_zero}.
% Additional results under various pruning ratios are presented in Appendix~\ref{apx:additional_results_perplexity}.

\subsection{Generalization Ability Evaluation}
We also compare the generalization ability of our reconstruction framework with existing approaches. Specifically, we validate the generalization ability with different pruning modules and criteria across diverse especially high pruning ratios. Here we adopt Naive pruning, Bias Compensation~\cite{an2023fluctuation}, and Mask-Tuning~\cite{kwon2022fast} as baselines, in which naive pruning signifies no reconstruction.

% We also compare our reconstruction framework with other methods under various conditions, like their suitability for different modules and their performance based on different pruning criteria. In our experiments, we use three recovery methods from previous studies as our benchmarks: naive pruning, Bias Compensation, and Mask Tuning. Naive pruning signifies no recovery, Bias Compensation is introduced by FLAP, and Mask Tuning is the recovery method used by \citeauthor{kwon2022fast}, which KCM also adopts.
\paragraph{Generalization to Pruning Modules.}

To assess the reconstruction capability of our framework on different modules, we apply LIAR to reconstruct attention heads and FFN hidden channels, which we refer to as the FFN neurons. Specifically, we first utilize the pruning criterion of \cite{kwon2022fast} to prune only heads or neurons of BERT$_{\rm BASE}$, then we apply different reconstruction algorithms to regain the performance, while the remaining elements stay frozen. 

\begin{figure*}[t]
    \centering
    \includegraphics[width=\linewidth]{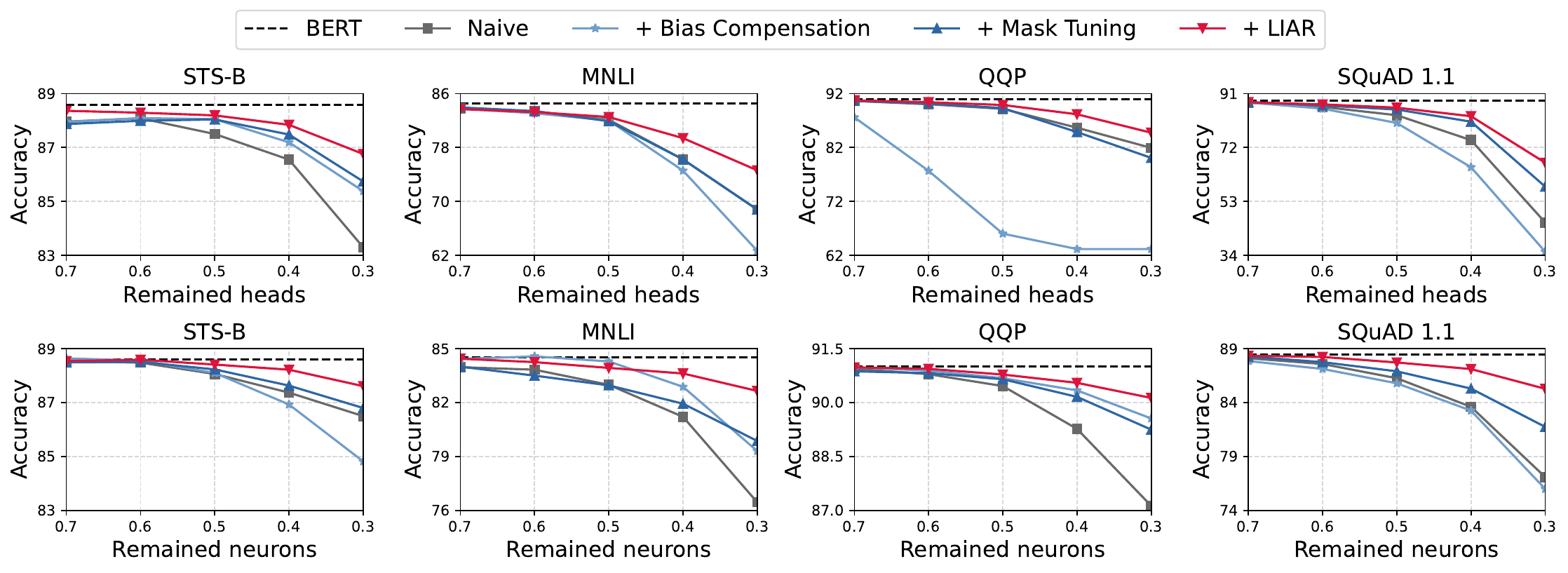}
    % \vspace{-3mm}
    \caption{Accuracy comparison of BERT$_{\rm BASE}$ on STS-B, MNLI, QQP and SQuAD$_{1.1}$ tasks by pruning attention heads (\textbf{upper}) and FFN neurons (\textbf{lower}) with different reconstruction strategies.
    % The dashed black line represents the accuracy of the BERT$_{\rm BASE}$, and the grey, blue, purple, and orange lines are the performance without pruning and with bias compensation, Mask-Tuning and LIAR respectively.
    }
    \label{fig:compatibility_module}
\end{figure*}

%As Figure~\ref{fig:compatibility_module} illustrates, Bias Compensation results in fairly unstable performance, particularly on the STS-B and QQP tasks. These tasks even show lower accuracy compared to naive pruning, likely due to significant fluctuations in the hidden states for some tasks and the inability of the estimated bias to reconstruct the distorted output. In contrast, LIAR and Mask Tuning demonstrate more consistent recovery gains. Notably, LIAR almost always significantly improves accuracy, especially when pruning most of the heads or neurons, e.g., 30\% and 10\%.
Figure~\ref{fig:compatibility_module} depicts that Bias Compensation leads to somewhat unstable performance, especially in the STS-B and QQP tasks, where it often results in lower accuracy than naive pruning. This decrease in accuracy might stem from substantial variations in the hidden states across certain tasks and the failure of the estimated bias to restore the altered outputs. In contrast, LIAR and Mask-Tuning show more stable improvements in recovery. Particularly, LIAR consistently enhances accuracy, markedly when a significant proportion of heads or neurons are pruned, such as 30\% and 10\%.

\begin{figure*}[t]
    \centering
    \includegraphics[width=\linewidth]{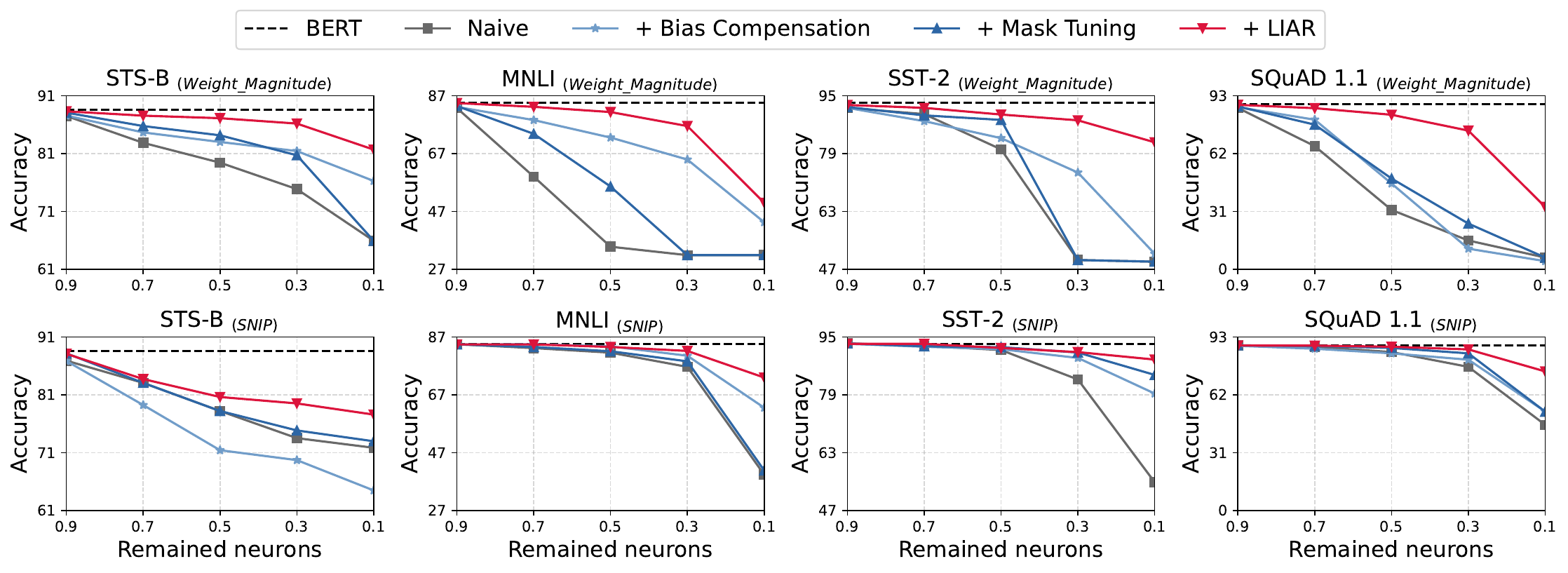}
    % \vspace{-3mm}
    \caption{Accuracy comparison of the BERT$_{\rm BASE}$ pruned by Weight Magnitude-based (\textbf{upper}) and SNIP (\textbf{lower}) criteria on the STS-B, MNLI, SST-2 and SQuAD$_{1.1}$ tasks respectively. We only prune the FFN neurons to avoid introducing the architecture search problem.}
    \label{fig:compatibility_metric}
\end{figure*}

\paragraph{Generalization to Pruning Criteria.}
\label{sec:compatibility_criteria}

Finally, to verify the generalization ability to manifold pruning algorithms, we apply LIAR on two retraining-based pruning criteria: Weight Magnitude~\cite{li2017pruning} and SNIP~\cite{lee2018snip}, which are zero-order and first-order respectively. 

Figure~\ref{fig:compatibility_metric} shows that LIAR significantly improves performance under all pruning ratios compared to simple pruning without any reconstruction. Furthermore, LIAR reduces the gap between different pruning metrics, for instance, on the SST-2 task. Interestingly, the STS-B accuracy derived from the Weight Magnitude-based criterion with 90\% neurons pruned outperforms SNIP after being reconstructed by LIAR, even though it performs worse without reconstruction. This likely indicates that less effective pruning metrics can have higher recoverability. More results about the generalization performance on other tasks are demonstrated in Appendix~\ref{apx:additional_results_module} and \ref{apx:additional_results_criteria}.

% \paragraph{Compatibility with Extreme Pruning Ratios}

% It is reasonable and almost inevitable that the model capabilities degrade catastrophically along with extremely high pruning ratios, in which case the model output is distorted too seriously to remain in the original distribution, especially for the retraining-free circumstance. Therefore, we believe it is a major issue to examine the performance of a recovery framework under extreme conditions to ensure stability and usability. We validate the 
% classification and QA performance of BERT$_{\rm BASE}$ under 80\% and 90\% pruning ratios respectively. The pruning criterion is weight magnitude-based.

% As Table xx elaborates, LIAR not only surpasses naive pruning on a large scale on all tasks substantially but also retains higher accuracy compared to Mask-Tuning, which confirms that LIAR is fairly practicable even under serious conditions.

\subsection{Ablation Study}
\label{sec:ablation}
%In this subsection, we (1) provide more visualization of the reconstruction error across the whole model, (2) analyze the efficacy of the updation for weight and bias terms of LIAR,
%(3) evaluate the robustness of LIAR in terms of calibration samples, and (4) assess the time consumption.
In this subsection, we will present additional visualizations of the reconstruction error throughout the entire model, analyze the effectiveness of updates to the weight and bias terms in LIAR, evaluate the robustness of LIAR with respect to calibration samples, and assess the time consumption involved.

\paragraph{Reconstruction Error.}

\begin{figure*}[t]
\centering
    \subfloat[]{
    \centering
    \includegraphics[width=0.49\linewidth]{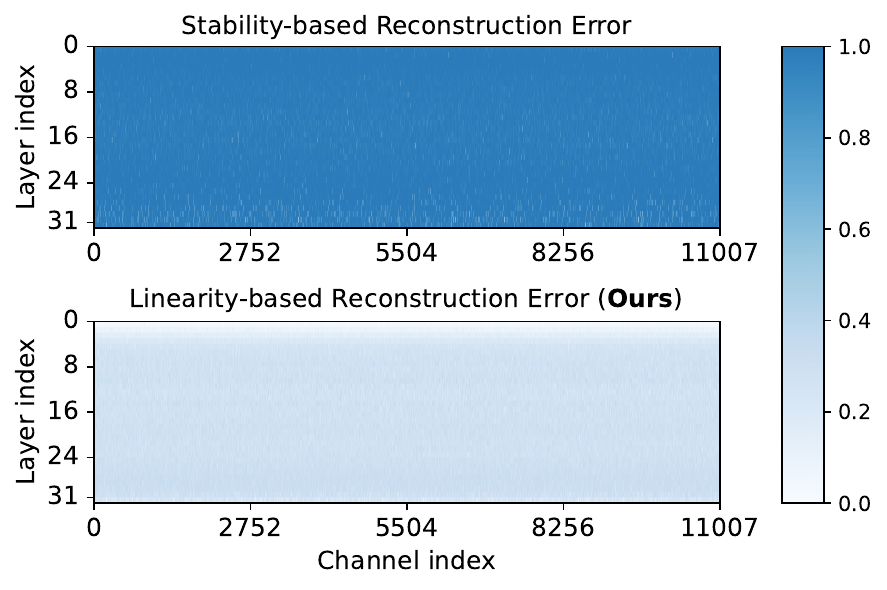}
    \label{fig:error_model}
    }
    \subfloat[]{
    \centering
    \includegraphics[width=0.49\linewidth]{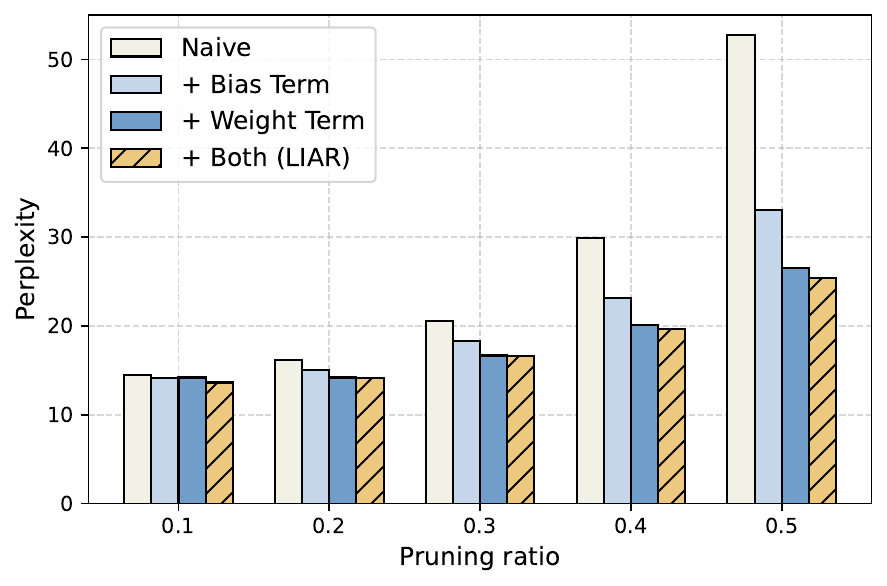}
    \label{fig:ablation_stability_linearity}
    }
    \vspace{-2mm}
    \caption{(a) Reconstruction error distribution of the hidden input of LLaMA-7B across 1024 instances sampled from the WikiText-2 traning dataset. (b) Perplexity comparison of the LLaMA-7B by removing the updation for the bias and weight term at various pruning ratios.
    % The light blue bars represent the performance derived by naive pruning without any recovery, while the dark blue and grey bars denote the perplexity recovered by only utilizing the stability and linearity respectively, and the orange bars are acquired by employing them simultaneously, which is also our proposed LIAR framework.
    }
\end{figure*}

While Figure~\ref{fig:single_channel} only depicted the reconstruction error for a single channel, we now present a more comprehensive analysis of the error distribution across the entire model. As demonstrated in Figure~\ref{fig:error_model}, LIAR significantly reduces the error in reconstructing the model's output.

% \begin{figure}[t]
% % \begin{wrapfigure}{l}{0.5\textwidth}
%     \centering
%     % \includegraphics[width=\linewidth, height=0.6\linewidth]{example-image-duck}
%     \includegraphics[width=0.5\linewidth]{fig/ablation_stability_linearity.pdf}
%     \vspace{-3mm}
%     \caption{Perplexity comparison of the LLaMA-7B with different recovery strategies at various pruning ratios. The light blue bars represent the performance derived by naive pruning without any recovery, while the dark blue and grey bars denote the perplexity recovered by only utilizing the stability and linearity respectively, and the orange bars are acquired by employing them simultaneously, which is also our proposed LIAR framework.}
%     \label{fig:ablation_stability_linearity}
% % \end{wrapfigure}
% \end{figure}

\paragraph{Interpolated Weight \& Bias.}

As mentioned earlier, we reconstruct the distorted output by integrating and updating a weight and bias term into the original matrices. To evaluate their significance, we modify Equation~(\ref{eq:final_equation}) by retaining the update for either the weight or the bias term, while omitting the other. The perplexity of the LLaMA-7B model, as illustrated in Figure~\ref{fig:ablation_stability_linearity}, demonstrates that both the weight and bias updates are vital for reconstructing the pruned components, with the weight term playing a more significant role.

\paragraph{Robustness to Calibration Samples.}

\begin{figure}
    \centering
    \begin{minipage}{.46\linewidth}
        \includegraphics[width=\linewidth]{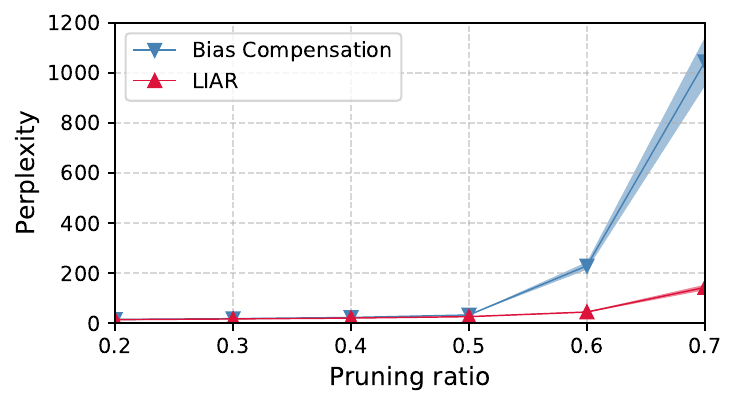}
        \caption{Dynamic perplexity range of LLaMA-7B on WikiText, which is fed with 128, 256, 512, and 1024 samples respectively.}
        \label{fig:ablation_batch_size}
    \end{minipage}
    \begin{minipage}{.53\linewidth}
        % \begin{table}[t]
            \centering
            \tabcaption{Time consumption (minutes) for reconstructing LLaMA models with different numbers of calibration samples. We conduct the experiments on one single NVIDIA Tesla A100 80G GPU.}
            % \resizebox{\linewidth}{!}{
                \begin{tabular}{l  | c  c c c}
                    \toprule
                    \multirow{2}{*}{Model} &  \multicolumn{4}{c}{Calibration Samples} \\
                    \cmidrule{2-5}
                    &  128 & 256 & 512 & 1024 \\
                    \midrule
                    LLaMA-7B & \textbf{1.0} & 1.5 & 2.1 & 3.9 \\
                    LLaMA-13B & 3.1 & 3.8 & 4.9 & 7.9 \\
                    LLaMA-30B & 6.6 & 8.5 & 12.7 & 19.7 \\
                    
                    \bottomrule
        
                \end{tabular}
            % }
            \label{tab:ablation_time}
        % \end{table}
    \end{minipage}
\end{figure}

Since we utilize a calibration dataset to assist in estimating the weight and bias terms of Equations~(\ref{eq:Xm=XuQ+Xm})-(\ref{eq:Wm=PWu+Wm}), we also evaluate the impact of the size of this dataset to ensure the usability under low-resource conditions. Specifically, we prune the LLaMA-7B model using FLAP, and apply Bias Compensation and LIAR to reconstruct it respectively. As shown in Figure~\ref{fig:ablation_batch_size}, LIAR consistently maintains a low dynamic perplexity range, consistently outperforming FLAP as the pruning ratio varies from 0.2 to 0.7. This indicates that our method is highly efficient, requiring only a few or even a single forward propagation, without the need for back-propagation.

% \begin{table}[t]
%     \centering
%     \caption{The comparison of the trade-off between the pruning time consumption and the perplexity on the WikiText-2 dataset by pruning 50\% parameters of LLaMA-7B. We conduct the experiments on one single NVIDIA Tesla A100 80G GPU. We compare the performance of FLAP addition with our LIAR framework against the prior structured pruning frameworks, where $^\ast$ denotes the results from \cite{an2023fluctuation}.}
%     % \resizebox{\linewidth}{!}{
%         \begin{tabular}{l  c  c  c}
%             \toprule
%             Method & Performance Recovery& Perplexity & Time (minutes) \\
%             \midrule
%             LLM-Pruner & Retrain & 38.12 & 120 \\
%             LLM-Pruner & Naive & 112.44 & 1 \\
%             FLAP & Naive & 33.00 & 2 \\
%             FLAP & Bias Compensation & 33.00 & 2 \\
%             \midrule
%             \rowcolor{orange!30} FLAP + LIAR & LIAR & 25.43 & 2\\
%             \bottomrule

%         \end{tabular}
%     % }
    
%     \label{tab:ablation_time}
% \end{table}

\paragraph{Time Consumption.}

To further analyze the efficiency of our approach, we record the time consumption for LLaMA models using varying numbers of calibration samples. Table~\ref{tab:ablation_time} shows that the time required increases with both model size and the number of samples. Notably, LIAR consistently demonstrates low time costs on larger models, particularly completing tasks with the LLaMA-7B model in just under one minute. It's important to highlight that LIAR was tested using a single GPU in these experiments, suggesting that its efficiency could improve even further with additional computational resources.

\section{Conclusion}
In this paper, we propose LIAR (\textbf{L}inear \textbf{I}nterpolation-based \textbf{A}daptive \textbf{R}econstruction), an efficient and effective reconstruction framework with no need for back-propagation and retraining and compatible with various pruning modules and criteria. LIAR leverages the preserved modules to approximate the masked ones, reconstructing the output distortion by applying linear interpolation to the preserved weight matrix. We empirically evaluate the validity of LIAR on GLUE, SQuAD, WikiText-2, and common sense reasoning benchmarks respectively, where the method with LIAR reduces 50\% encoder parameters within only 2\% accuracy degradation for BERT$_{\rm BASE}$, and achieves 2.56$\times$ performance enhancement for LLaMA-7B under the 50\% pruning ratio within 1 minute.
% Our extensive experiments also illustrate that LIAR is more universally compatible with various modules and pruning metrics under wide-ranged pruning ratios compared to existing recovery algorithms.

% \begin{ack}

% \end{ack}

% \section*{References}

% References follow the acknowledgments in the camera-ready paper. Use unnumbered first-level heading for
% the references. Any choice of citation style is acceptable as long as you are
% consistent. It is permissible to reduce the font size to \verb+small+ (9 point)
% when listing the references.
% Note that the Reference section does not count towards the page limit.
% \medskip

\normalem
% \bibliography{reference.bib}
% \input{neurips_2024.bbl}
% \bibliography{reference}

%%%%%%%%%%%%%%%%%%%%%%%%%%%%%%%%%%%%%%%%%%%%%%%%%%%%%%%%%%%%

% \newpage
\appendix

% \section{Appendix / supplemental material}

\appendix
\onecolumn

\section{Implementation Algorithm of LIAR}
\label{apx:algorithm}
The detailed steps of our method are outlined in Algorithm \ref{alg:liar}. The inputs to LIAR encompass the pre-trained language model $\mathcal{M}$ with $L$ attention and FFN layers, each of which contains weight $\mathbf{W}^\ell$, bias $\mathbf{B}^\ell$ and mask variables $\mathbf{M}^\ell$. We conduct the reconstruction based on the calibration dataset $\mathcal{D}$, and return a well-pruned and -recovered model $\mathcal{S}$.

Our approach decomposes the reconstruction problem for the model $\mathcal{M}$ into layer-wise subproblems. Specifically, we first collect the hidden input for the first layer, then iteratively solve the subproblem and derive the input for the next one. For each subproblem, we first split the input and weight into the unmasked and masked ones respectively based on the mask variables. Building on this, we calculate the mean value for the input to derive the stable pattern given varying samples, and then estimate the transformation matrices for the masked input and weight by solving the least-square problem. Based on the estimated correlations between the pruned and preserved components, the weight and bias are updated accordingly.

\begin{algorithm}[htbp]
\renewcommand{\algorithmicrequire}{\textbf{Input:}}
\renewcommand{\algorithmicensure}{\textbf{Output:}}

\caption{Linear Interpolation-based Adaptive Reconstruction (LIAR) Framework.}
\begin{algorithmic}[1]
\Require Pre-trained language model $\mathcal{M}$ with $L$ layers. Mask variables $\mathbf{M}^\ell$, weight $\mathbf{W}^\ell$, and bias $\mathbf{B}^\ell$ for layer $\ell$. Calibration dataset $\mathcal{D}$.
\State Initialize model $\mathcal{S}$: $\mathcal{S} \leftarrow \mathcal{M}$.
\For {sample in $\mathcal{D}$}
\State Collect hidden input $\mathbf{X}^\ell$ for $\ell=1$ of model $\mathcal{S}$.
\EndFor
\For{$\ell\leftarrow 1$ to $L$}
\State Split $\mathbf{X}^\ell$ and $\mathbf{W}^\ell$ into $\mathbf{X}_u^\ell$, $\mathbf{X}_m^\ell$, $\mathbf{W}_u^\ell$, and $\mathbf{W}_m^\ell$ respectively based on the mask $\mathbf{M}^\ell$.
\State Calculate the mean value of $\mathbf{X}^\ell$: $\overline{\mathbf{X}}^{\ell}=\frac{1}{NT} \sum\limits_{i=1}^{N} \sum\limits_{j=1}^{T} \mathbf{X}^\ell_{i,j,:}$.
\State Estimate $\mathbf{Q}^\ell$ for the input: $\mathbf{Q}^\ell = \mathop{\arg\min}\limits_{\mathbf{\mathbf{Q}^\ell}}  \left\Vert \mathbf{X}^\ell_m - \left( \mathbf{X}^\ell_u \mathbf{Q}^\ell + \overline{\mathbf{X}}^\ell_m \right) \right\Vert^2_2$.
\State Estimate the transformation matrix $\mathbf{P}^\ell$ for the weight: $\mathbf{P}^\ell = \text{argmin}_{\mathbf{\mathbf{P}^\ell}}  \left\Vert \mathbf{W}^\ell_m -  \mathbf{P}^\ell \mathbf{W}^\ell_u  \right\Vert^2_2$.
\State Update the weight: $\mathbf{W}^\ell \leftarrow (\mathbf{I} + \mathbf{Q}^\ell \mathbf{P}^\ell)\mathbf{W}^\ell_u$.
\State  Update the bias: $\mathbf{B}^\ell \leftarrow \overline{\mathbf{X}}^\ell_m \mathbf{W}^\ell_m + \mathbf{B}^\ell$.
\State Collect hidden input $\mathbf{X}^{\ell+1}$ based on the updated weight $\mathbf{W}^\ell$ and bias $\mathbf{B}^\ell$.
\EndFor
\Ensure{Pruned and recovered model $\mathcal{S}$}

\end{algorithmic}
\label{alg:liar}
\end{algorithm}

\section{Experimental Setup}
\label{apx:experimental_setup}

\subsection{Models}
We introduce two categorized models in our experiments: BERT$_{\rm BASE}$ \cite{devlin2018bert} and LLaMA family models \cite{touvron2023llama}. BERT$_{\rm BASE}$ of 0.1B parameters is an encoder-based model, which is stacked by 12 Transformer layers, each of which has 12 attention heads and 3072 FFN neurons, and the embedding dimension is 768. LLaMA is a set of decoder-based large language models open-sourced by Meta, mainly including LLaMA-7B/13B/30B/65B, and limited by the computing resources, we only choose the 7B, 13B, and 30B sizes to conduct experiments. Take the LLaMA-7B as an example, it consists of 32 decoder layers, whose embedding size is 4096, and the number of attention heads and FFN neurons are 32 and 11008 respectively.

\subsection{Tasks}
\label{apx:task}

\paragraph{GLUE Benchmark.} The GLUE (General Language Understanding Evaluation) benchmark is a collection of datasets for evaluating the performance of models across a diverse set of existing natural language understanding tasks. GLUE consists of 3 categorized sequence classification tasks: 1) Natural language inference (MNLI \cite{williams2017broad}, QNLI \cite{tambe2021edgebert}, WNLI \cite{levesque2012winograd}, RTE \cite{dagan2005pascal,haim2006second,giampiccolo2007third,bentivogli2009fifth} with 393K, 105K, 0.6K and 2.5 K training samples), 2) Single-sentence classification (SST-2 \cite{socher2013recursive}, CoLA \cite{warstadt2019neural} with 67K and 8.5K training examples), and 3) Similarity and paraphrase (QQP \cite{qqp2017}, MRPC \cite{dolan2005automatically}, STS-B \cite{cer2017semeval} with 364K, 3.7K and 7K training samples respectively). We exclude the WNLI, RTE, and CoLA tasks due to their unstable performance. For the other tasks, we prune the model by using the training set of each dataset for different tasks and report the Accuracy on the development sets except for the STS-B task, for which we report the Spearman Correlation.

\paragraph{SQuAD.} The SQuAD (Stanford Question Answering Dataset) is a reading comprehension dataset for the question-answering task, which is categorized into 2 versions: SQuAD$_{1.1}$ \cite{rajpurkar2016squad} and SQuAD$_{2.0}$ \cite{rajpurkar2018know}, each of which contains 88K and 130K training examples. To be specific, SQuAD$_{2.0}$ is an extension of SQuAD$_{1.1}$ by including unanswerable questions about the same paragraphs whose answers are not stated in the given contexts. We report the F1 score for SQuAD tasks.

\paragraph{WikiText.} The WikiText corpus is a language modeling benchmark dataset, which is a hundred times larger than the previous Penn Treebank \cite{marcus1993building}. This corpus is available in two data sizes: WikiText-2 and WikiText-103, which have 2K and 103K training samples respectively,  Both datasets use the same articles for validation and testing. We follow \cite{an2023fluctuation} to conduct pruning on the WikiText-2 training set and report the perplexity metric for the test set, which gauges a model’s predictive accuracy.

\paragraph{Common Sense Reasoning Benchmarks.} We conduct experiments on seven pivotal common sense benchmarks to evaluate the model's zero-shot performance: BoolQ \cite{clark2019boolq}, PIQA \cite{bisk2020piqa}, HellaSwag \cite{zellers2019hellaswag}, WinoGrande \cite{sakaguchi2021winogrande}, ARC-easy \cite{clark2018think}, ARC-challenge \cite{mihaylov2018can} and OpenbookQA \cite{mihaylov2018can}. We conduct the reconstruction on the Alpaca dataset \cite{alpaca} and report both the accuracy of each benchmark and the overall average accuracy.

\subsection{Baselines}
\paragraph{Retraining-free Pruning Methods.}

% \begin{table}[h]
%     \centering
%     \caption{Mapping}
%     \begin{tabular}{l | c c c c c c c}
%         \toprule
%          Paradigm & Method & Model & Pruning Criterion & Reconstruction \\
%          % \multirow{2}{*}{Method} & \multicolumn{7}{c}{Pruning Ratio} \\
%          % \cmidrule(lr){2-8}
%          \midrule
%          \multirow{3}{*}{Retraining-free} & \cite{kwon2022fast} & BERT & Mask-Search & Mask-Tuning \\
%          & KCM~\cite{nova2023gradient} & BERT & R2D2 & Mask-Tuning \\
%          & FLAP~\cite{an2023fluctuation} & LLaMA, Vicuna & Fluctuation Metric & Bias Compensation \\
%          \midrule
%          \multirow{3}{*}{Retraining-based} & \cite{li2017pruning} & - & Weight Magnitude & - \\
%          & SNIP~\cite{lee2018snip} & - & - \\
%          & LLM-Pruner & LLaMA, Vicuna, ChatGLM \\

%          \bottomrule
%     \end{tabular}
    
%     \label{tab:apx_perplexity}
% \end{table}

To our best knowledge, \cite{kwon2022fast} is the first post-training pruning framework for transformers, which proposed Mask Search and Mask Rearrangement to measure the importance of attention heads and neurons based on the Fisher information, and Mask Tuning to recover its accuracy. KCM (Kernelized Convex Masking) \cite{nova2023gradient} proposed two ranking techniques to estimate the importance of individual neurons: Representative Ranking (R2) and Data-Driven (D2). It is noticed that KCM only prunes the neurons, and it brings about significant performance degradation under large compression rates as shown in Table~\ref{tab:exp_bert}. While previous studies concentrate on BERT-based models, FLAP (Fluctuation-based Adaptive Structured Pruning) focuses on decoder-based models such as the LLaMA model family and the Vicuna-7B model \cite{vicuna2023}. FLAP prunes the most stationary heads and neurons and utilizes the average value of the pruned components as a bias to compensate for the output error.

\paragraph{Existing Reconstruction Algorithms.} Naive pruning means direct pruning without any retraining or reconstruction, which represents the efficacy of the pruning criteria to some extent and serves as a baseline for different reconstruction strategies. Mask Tuning is a reconstruction technique proposed by \cite{kwon2022fast}, which rescales the nonzero mask values to any real values instead of being restricted to 1 by layer-wise reconstruction via linear least squares. This technique is also utilized by KCM. Bias Compensation is proposed by \cite{an2023fluctuation}, which operates by calculating the average value of the attention or FFN output matrix and multiplying it with the pruned weight to obtain a bias to compensate for the distorted output.

\subsection{Pruning Criteria.}
We introduce two pruning metrics that require retraining to avoid hurting the performance: Weight Magnitude and SNIP.

\paragraph{Weight Magnitude.} Weight Magnitude \cite{li2017pruning} is a conventional pruning criterion based on the magnitude of filters, which utilizes the zero-order information of the weight and prunes the filters with small magnitudes. The importance score for $k$-th filter $\mathcal{F}_k \in \mathbb{R}^{n_i \times l \times l}$ with $n_i$ input channels and kernel width $l$ is calculated as
\begin{equation}
    s_k=\sum\limits \left\Vert \mathcal{F}_k \right\Vert_1.
\end{equation}

\paragraph{SNIP.} SNIP (Single-shot Network Pruning) \cite{lee2018snip} is a simple but effective technique that measures the sensitivity of the connections and allows us to identify and prune the redundant connections in a single step. To be specific, given the dataset $\mathcal{D}$ and weight $\mathbf{w}$ with $m$ connections, the effect of removing connection $c_k$ is first calculated by
\begin{equation}
    \Delta L_k(\mathbf{w} ; \mathcal{D}) \approx g_k(\mathbf{w} ; \mathcal{D}) = \frac{\partial L (\mathbf{c} \odot \mathbf{w} ; \mathcal{D})}{\partial c_k} \bigg|_\mathbf{c=1},
\end{equation}
where $\mathbf{c}\in\{0,1\}^m$ are indicator variables representing the connectivity of $\mathbf{w}$. The magnitude of the derivatives $g_k$ is taken as the saliency criterion and normalized to obtain the importance score:
\begin{equation}
    s_k = \frac{|g_k(\mathbf{w} ; \mathcal{D})|}{\sum_{j=1}^{m}|g_j(\mathbf{w} ; \mathcal{D})|}.
\end{equation}
The connections with the least importance scores are removed directly accordingly.

\section{Detailed Implementation}
\label{apx:detailed_implementation}

\subsection{Tasks \& Datasets}
We download the datasets from the Huggingface repositories on GLUE, SQuAD, WikiText, BoolQ, PIQA, HellaSwag, WinoGrande, ARC, and OBQA benchmarks. We employ the EleutherAI LM Harness\footnote{\url{https://github.com/EleutherAI/lm-evaluation-harness}}, a public evaluation benchmark, to evaluate the zero-shot performance on the seven common sense reasoning benchmarks.
% GLUE\footnote{\url{https://huggingface.co/datasets/glue}}, SQuAD\footnote{\url{https://huggingface.co/datasets/squad}}, WikiText\footnote{\url{https://huggingface.co/datasets/wikitext}}, BoolQ\footnote{\url{https://huggingface.co/datasets/boolq}}, PIQA \footnote{\url{https://huggingface.co/datasets/piqa}}, HellaSwag\footnote{\url{https://huggingface.co/datasets/hellaswag}}, WinoGrande\footnote{\url{https://huggingface.co/datasets/winogrande}}, ARC\footnote{\url{https://huggingface.co/datasets/ai2_arc}}, and OBQA\footnote{\url{https://huggingface.co/datasets/openbookqa}} benchmarks.

\subsection{Models}
We download the BERT$_{\rm BASE}$ model\footnote{\url{https://huggingface.co/bert-base-uncased}} and LLaMA-7B model\footnote{\url{https://huggingface.co/linhvu/decapoda-research-llama-7b-hf}} from the Huggingface repository. To obtain the task-specific BERT model, we finetune the pre-trained BERT$_{\rm BASE}$ to the downstream tasks following a standard training recipe.

\subsection{Implementation for Retraining-free Pruning Methods}

We utilize the released code by authors\footnote{\url{https://github.com/WoosukKwon/retraining-free-pruning}} to implement \cite{kwon2022fast}. We use damp=1 for LSMR solver\footnote{cupyx.scipy.sparse.linalg.lsmr} in CuPy and an acceptable range of tuned variables as $[-10, 10]$ as the paper described. All of the experimental settings are kept to default. As for the KCM, we reimplement it since there is no public implementation of authors. We use width $\sigma=1$ for the Gaussian kernel and convergence rate $\alpha=0.01$ in the paper \cite{nova2023gradient}. We use Z-score normalization for normalizing the D2 scores. Towards the FLAP, we follow of implementation of authors\footnote{\url{https://github.com/CASIA-IVA-Lab/FLAP}}.

\subsection{Implementation for Reconstruction Algorithms}
As part of \cite{kwon2022fast} and FLAP, we implement Mask Tuning and Bias Compensation following the released code and fix the samples to be 2048, 1024, and 1024 for Mask Tuning, Bias Compensation, and LIAR respectively.

\subsection{Implementation for Pruning Criteria}

\paragraph{Weight Magnitude.} As the weight magnitude-based pruning method is originally designed for convolutional kernels, which is not well-suited for pruning FFN neurons of the transformer block, we modify the metric to align with the characteristics of the transformer layer. We also utilize the $\ell_2$-error as it yields higher accuracy according to our experimental results, and the resulting accuracy is similar to the implementation in \cite{nova2023gradient}. The adapted metric takes the following term:
\begin{equation}
    s_k=\sum\limits_{i=1}^{C_{out}} \left\Vert \mathbf{W}_{k,i} \right\Vert^2_2,
\end{equation}
where $\mathbf{W} \in \mathbb{R}^{C_{in} \times C_{out}}$ is the output matrix of the FFN layers.

\paragraph{SNIP.} The size of dataset $\mathcal{D}$ for pruning is fixed to 5K for all tasks, which yields fairly stable performance according to our empirical analysis.

\subsection{Other Details}
We implement our framework and the baselines with Pytorch \cite{paszke2019pytorch} and Huggingface Transformers \cite{wolf2020transformers} libraries. To save the memory consumption, we load the model onto the GPU in 16-bit floating-point format when pruning the LLaMA models.

\subsection{Generalization to Pruning Modules on Other Tasks}
\label{apx:additional_results_module}

\begin{figure*}[h]
    \centering
    \includegraphics[width=\linewidth]{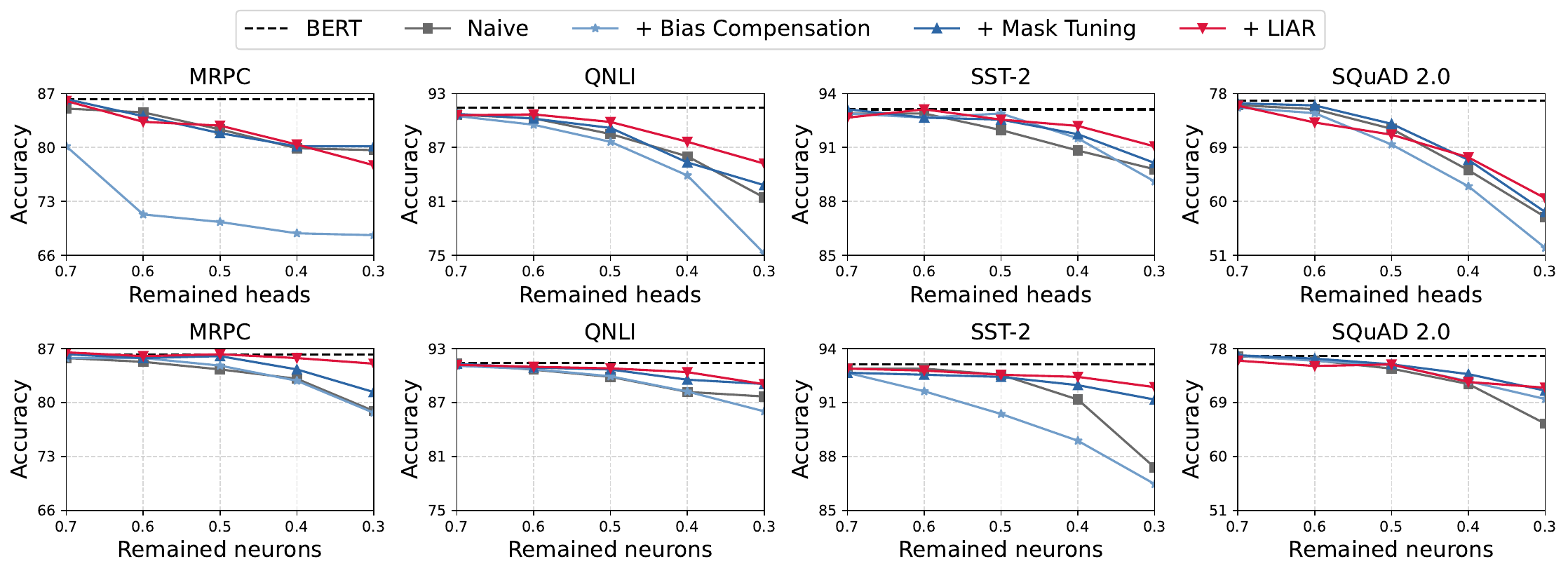}
    \caption{Performance comparison of the BERT$_{\rm BASE}$ on MRPC, QNLI, SST-2 and SQuAD$_{2.0}$ tasks by pruning attention heads (\textbf{upper}) and FFN neurons (\textbf{lower}) with different reconstruction strategies.}
    \label{fig:appendix_compatibility_module}
\end{figure*}

Figure~\ref{fig:appendix_compatibility_module} demonstrates the performance comparison on MRPC, QNLI, SST-2, and SQuAD$_{2.0}$ tasks by removing attention heads and FFN neurons respectively based on the importance score derived by \cite{kwon2022fast}. Our approach recovers the pruned output and achieves higher accuracy on most occasions.

\subsection{Generalization to Pruning Criteria on Other Tasks}
\label{apx:additional_results_criteria}

\begin{figure*}[h]
    \centering
    \includegraphics[width=\linewidth]{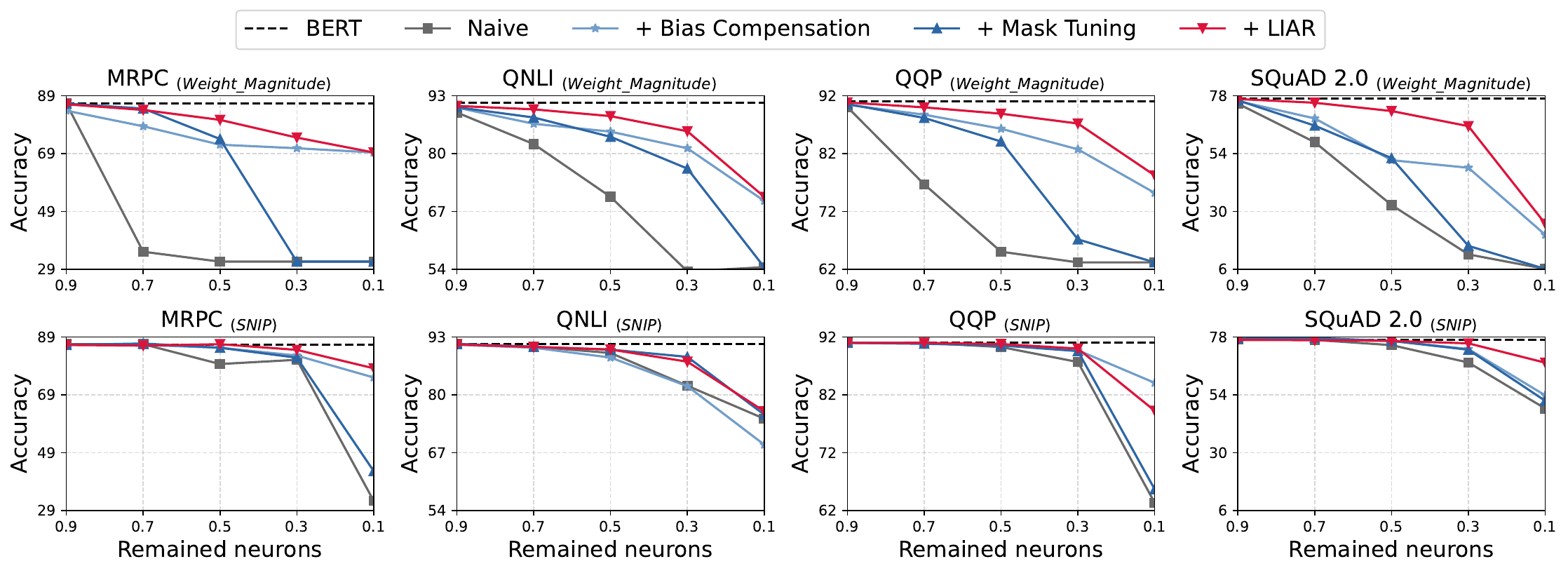}
    \caption{Performance comparison of the BERT$_{\rm BASE}$ pruned by Weight Magnitude-based (\textbf{upper} lines) and SNIP (\textbf{lower} lines) criteria on the MRPC, QNLI, QQP and SQuAD$_{2.0}$ tasks respectively, in which we only prune the FFN neurons to avoid to introduce the architecture search problem.}
    \label{fig:appendix_compatibility_metric}
\end{figure*}

Figure~\ref{fig:appendix_compatibility_metric} shows the effectiveness of our method for two retraining-based pruning criteria: Weight Magnitude and SNIP. It is obvious that LIAR attains the most consistent and significant performance improvements across both pruning metrics and nearly all the tasks and pruning ratios.

\section{Limitations}
\label{apx:limitation}
Although this study brings significant performance enhancement for retraining-free pruning, it still faces two important potential limitations for future research directions. (1) Firstly, as our method utilizes calibration samples, it shares a similar and common issue with all of the data-driven approaches, which will encounter an overfitting problem with limited data. Most works choose to facilitate it by feeding numerous samples to guarantee stable performance (e.g., retraining-based methods) while our approach has a rather lower requirement for the dataset size compared to conventional data-driven approaches as demonstrated in Figure~\ref{fig:ablation_batch_size} and thus to be more efficient. (2) Secondly, as our method is applicable to varied pruning metrics and does not involve determining the network architecture, the regained performance is dependent on the quality of the pruning criteria. In other words, whether a pruned model can be reconstructed through LIAR is based on whether the pruned parts are recoverable. Or, to put it another way, we think this characteristic may not be a drawback, but exactly makes our method a powerful validation tool to evaluate the quality of a specific pruning criterion.

\section{Broader Impacts}
\label{apx:broader_impacts}
In this paper, we introduce a method that stands out for its computational efficiency and the elimination of the need for retraining, while still delivering enhanced performance metrics. Our innovation paves the way for rapid compression and deployment processes for large language models, making it an invaluable resource for scenarios constrained by limited computational capabilities. Through comprehensive analysis, we have yet to identify any adverse effects associated with our proposed method, underscoring its potential for widespread application without negative repercussions.

\end{document}